\documentclass{article} % For LaTeX2e
\usepackage{iclr2025_conference,times}

\usepackage{amsmath,amsfonts,bm}

\def\eqref#1{equation~\ref{#1}}
\def\1{\bm{1}}

\DeclareMathAlphabet{\mathsfit}{\encodingdefault}{\sfdefault}{m}{sl}
\SetMathAlphabet{\mathsfit}{bold}{\encodingdefault}{\sfdefault}{bx}{n}

\usepackage{hyperref}
\usepackage{url}
\usepackage{graphicx}
\usepackage{booktabs}
\usepackage{subcaption}
\usepackage{svg}
\usepackage{enumitem}
\usepackage{arydshln}
\usepackage{wrapfig}

\definecolor{darkgreen}{rgb}{0.0, 0.5, 0.0}

\title{Assessing Open-World Forgetting in \\ Generative Image Model Customization}
\author{\qquad\qquad\qquad\qquad\quad Héctor Laria \quad Alex Gomez-Villa \quad Kai Wang \\
\qquad\qquad\qquad\qquad\qquad \textbf{Bogdan Raducanu} \quad \textbf{Joost van de Weijer} \\
\qquad\qquad\qquad Computer Vision Center, Universitat Autònoma de Barcelona
}

\iclrfinalcopy % Uncomment for camera-ready version, but NOT for submission.
\begin{document}

\maketitle

\begin{figure}[h]
\includegraphics[width=\textwidth]{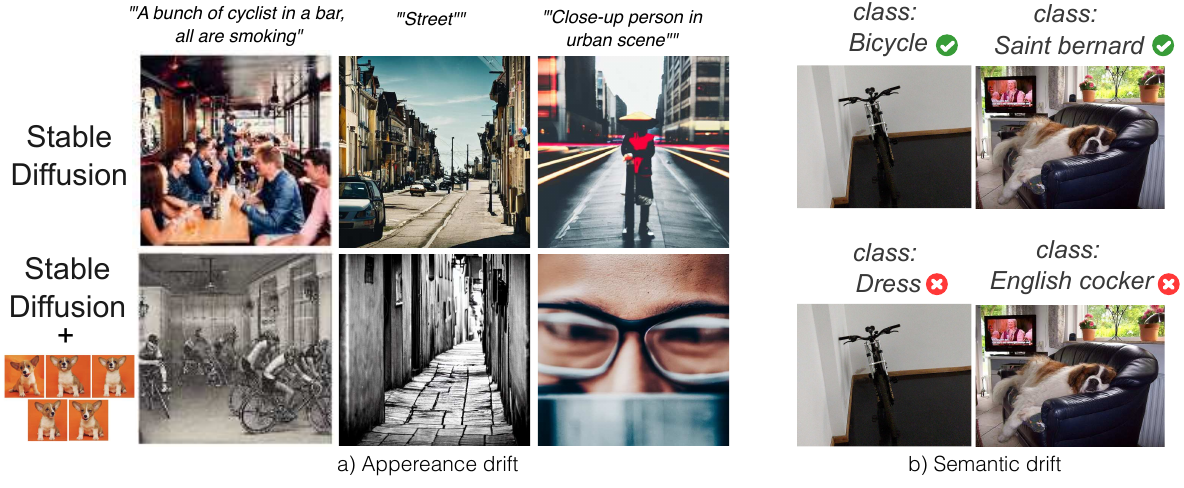}
\vspace{-1mm}
\caption{
\textbf{Unintended consequences in diffusion model customization}. Methods like Dreambooth lead to substantial drift in previously learned representations during the finetuning process even when adapting to as few as five images: a) Appearance drift:  Columns demonstrate  fine-grained class changes, complete object and scene shifts, and alterations in color (on both rows, images are generated from same seed). b) Semantic drift: finetuning negatively impacts the zero-shot classification capabilities of the models.
}
\label{fig:teaser}
\end{figure}

\begin{abstract}
Recent advances in diffusion models have significantly enhanced image generation capabilities. However, customizing these models with new classes often leads to unintended consequences that compromise their reliability. 
We introduce the concept of \textit{open-world forgetting} to characterize the vast scope of these unintended alterations.
Our work presents the first systematic investigation into open-world forgetting in diffusion models, focusing on semantic and appearance drift of representations. 
Using zero-shot classification, we demonstrate that even minor model adaptations can lead to significant semantic drift affecting areas far beyond newly introduced concepts, with accuracy drops of up to 60\% on previously learned concepts. Our analysis of appearance drift reveals substantial changes in texture and color distributions of generated content.
To address these issues, we propose a functional regularization strategy that effectively preserves original capabilities while accommodating new concepts. 
Through extensive experiments across multiple datasets and evaluation metrics, we demonstrate that our approach significantly reduces both semantic and appearance drift.
Our study highlights the importance of considering open-world forgetting in future research on model customization and finetuning methods.
\end{abstract}

\section{Introduction}
\label{sec:intro}

Given the high-quality image generation capabilities of recent models~\citep{deepfloyd,ramesh2022dalle2}, a major focus of research has been on how to efficiently incorporate new content and adapt them to new tasks and domains. To address this challenge, state-of-the-art transfer learning methods have been introduced, including: finetuning approaches like DreamBooth~\citep{ruiz2023dreambooth} and CustomDiffusion~\citep{kumari2022customdiffusion} for learning new concepts, conditioning-based methods like ControlNet~\citep{zhang2023controlnet} and IP-Adapters~\citep{ye2023ip-adapter} for precise generation control, prompt-based techniques for semantic editing, and parameter-efficient methods like low-Rank Adaptations(LoRA)~\citep{hu2022lora} for rapid adaptation with minimal overhead. These techniques enable models to learn new concepts effectively, even with few examples.

Finetuning approaches for adapting diffusion models~\citep{ruiz2023dreambooth, kumari2022customdiffusion} mainly rely on transfer learning and primarily focus on finetuning model weights to accommodate newly introduced data. This weight update can have unintended consequences, altering the model's existing knowledge and skills, or the alignment between language and visual content within the network, a phenomenon known as \emph{catastrophic forgetting} in continual learning research~\citep{kirkpatrick2017overcoming,de2021continual}. 
Traditionally, this phenomenon has been studied on what we term \emph{closed-world forgetting}, where evaluation is limited to a fixed set of classes encountered in previously learned tasks or skills. This setting assumes a clear, predefined set of concepts to evaluate against. In contrast, modern foundation models introduce what we term \emph{open-world forgetting}: degradation of the model's capabilities across its vast, unconstrained knowledge space. Unlike closed-world settings, open-world forgetting is particularly challenging to measure since the model's prior knowledge spans countless concepts, making it impossible to exhaustively evaluate what has been forgotten or altered during the adaptation process.

In this paper, we focus on finetuning-based personalization methods
for a case study of open-world forgetting. 
These techniques are especially interesting, as they only add little new knowledge to the network: a single new concept represented by a small set of typically 3-5 images. If open-world forgetting occurs even in this setting, where adaptation is limited to a handful of images, it strongly suggests that the phenomenon extends to larger-scale adaptations as well.
 Although one might expect that finetuning the model with such limited data would have minimal impact on the vast knowledge of the foundation model (e.g., Stable Diffusion), our analysis reveals that even these small updates can lead to highly detrimental consequences. As Figure~\ref{fig:teaser} illustrates, finetuning can drastically alter the image representation of concepts seemingly unrelated to the training images. The complexity of the forgetting underscores the need for a better understanding of how and where it occurs. Without this understanding, finetuned models risk becoming less reliable, less robust, and ultimately less trustworthy, particularly in safety-critical applications where precision and predictability are paramount.

We propose to analyze open-world forgetting from several perspectives. First, we examine \emph{semantic drift} using the recent observation that diffusion models can function as zero-shot classifiers; we propose to compare zero-shot capacity of models before and after adaptation on a set of image classification data sets. Second, we analyze \emph{appearance drift}  by evaluating changes in color and perceptual measurements before and after adaptation. Lastly, we assess the extent of forgetting in closely related concepts (\emph{local drift}) versus unrelated concepts. To address these three aspects of drift, we explore a straightforward, yet effective mitigation strategy by introducing a regularization technique during the training of new concepts. In conclusion, the main contributions of this work are:
\begin{itemize}[leftmargin=*]
    \item We are the first to systematically analyze \emph{open-world forgetting} in diffusion models due to model adaptation. Results show that even when adapting to very small domains, the consequences can be highly detrimental.  

    \item We propose two approaches to analyze \textit{open-world forgetting}, which are designed to assess \emph{semantic} and \emph{appearance drift} caused by the adaptation. We leverage the zero-shot classification capabilities of diffusion models to measure semantic drift, and observe drastic performance drops (of over 60\% for some classes). Appearance drift analysis confirms that customization leads to considerable changes in intra-class representation, color, and texture.

    \item We introduce a method to mitigate open-world forgetting, addressing the challenges of observed drift in text-to-image (T2I) models. 
    This method aims to preserve the original model's capabilities while allowing for effective customization. Experiments confirm that it greatly reduces both the semantic and appearance drift caused by open-world forgetting. 

\end{itemize}

\section{Related work}

\textbf{Text-to-image diffusion model adaptation.} Recent work has explored various approaches for personalizing text-to-image models. One line of research focuses on learning new concept tokens while freezing the generative backbone, pioneered by Textual Inversion~\cite{textual_inversion} and further developed in subsequent works~\cite{dong2022dreamartist,daras2022multiresolution,voynov2023ETI,hiper2023}. Another approach involves finetuning the model while updating modifier tokens, exemplified by DreamBooth and Custom Diffusion. On these, the pre-trained T2I model learns to bind a modified unique identifier to a specific subject given 3$\sim$5 images, while it also updates the T2I model parameters.

Finetuning methods often achieve state-of-the-art performance but introduce forgetting in large T2I models, as we will demonstrate in this article. While research focuses on improving new concept generation, it overlooks continuous model updating and forgetting mitigation. Recent works \citep{sun2024create_your_world,smith2023clora} address token forgetting but neglect other impacts of finetuning, such as semantic drifting in color, appearance, and visual recognition, which this paper explores.

It should be noted that next to the training-time adaptation methods described above, recently also inference-time customization methods have been developed. These methods~\citep{nair2023steered,brack2023sega,ruiz2024hyperdreambooth,zhou2024customization} achieve adaptation by modifying the generation process without parameter updates, typically through diffusion guidance or prompt engineering. These methods are not considered in our paper since they do not consider parameter updates of the model. 

\paragraph{Assessing forgetting.}
The vast majority of studies on \emph{Catastrophic forgetting} focus on, what we here call, \textit{closed-world forgetting}, where the knowledge of the network can be represented by its performance on a limited set of classes~\citep{lopez2017gradient,de2021continual, masana2022class}. However, as argued in the introduction, the growing importance of starting from large pretrained models (also known as foundation models), which have a vast prior knowledge, requires new techniques to assess forgetting. The forgetting of large language models (LLMs) during continual finetuning has received some attention in recent years, showing the importance of pretraining to mitigate forgetting~\citep{cossu2024continual}, however, they mainly evaluate on down-stream-task performance~\cite{scialom2022fine}.
To the best of our knowledge, \textit{open-world forgetting} has not yet been systematically analyzed for text-based image generation models, whose multi-modal nature can further worsen the impact of forgetting due to misalignment of the modalities. 

\section{Customization of Diffusion Models}
In this section, we briefly introduce T2I models and the two main customization methods we will evaluate during our analysis. In addition, we introduce an alternative regularization method to further mitigate forgetting. 

\subsection{Diffusion Models}
Diffusion models generate data by gradually denoising samples from a noise distribution through a forward and reverse process. The forward process adds Gaussian noise over $T$ steps via a Markov chain:

\begin{equation}
    q(\mathbf{x}_t | \mathbf{x}_{t-1}) = \mathcal{N}(\mathbf{x}_t; \sqrt{1 - \beta_t} \mathbf{x}_{t-1}, \beta_t \mathbf{I}),
\end{equation}

\noindent where $\beta_t$ controls the noise schedule. The reverse process learns to denoise by predicting the added noise through a neural network $\epsilon_\theta(x_t, t)$, optimizing:

\begin{equation}
    L(\theta) = \mathbb{E}_{t, \mathbf{x}_0, \epsilon} \left[ \left\| \epsilon - \epsilon_\theta(\mathbf{x}_t, t) \right\|^2 \right].
\end{equation}

Text-to-image diffusion models incorporate an additional conditioning vector $\mathbf{c} = \mathcal{E}(P)$ from a text encoder $\mathcal{E}$ and prompt $P$, enabling high-quality, diverse image generation conditioned on text descriptions.

\subsection{Customization Approaches}

Diffusion models often require finetuning for specific domains or user needs. This involves introducing new conditioning mechanisms or retraining on specialized datasets. This paper applies two adaptation methods to evaluate finetuning's impact on image generation models.

{\bf Dreambooth}~\citep{ruiz2023dreambooth} enables personalization of diffusion models by finetuning them with a small set of images. It reuses an infrequent token of the vocabulary to represent a unique subject, allowing the model to generate images of the subject in varied contexts or styles. This approach induces \textit{language drift} and \textit{reduced output diversity} in the model, which is mitigated by replaying class-specific instances alongside the subject training, called \textit{prior preservation loss}. The final training objective reads
\begin{equation} \label{eq:db_with_prior_reg}
    \mathbb{E}_{\mathbf{\epsilon}, \mathbf{x}, \mathbf{c}, t} [w_t \lVert \mathbf{\epsilon} - \mathbf{\epsilon}_\theta (\mathbf{x}_t, \mathbf{c}, t) \rVert + \lambda w_{t'} \lVert \mathbf{\epsilon}' - \mathbf{\epsilon}_\theta (\mathbf{x}_{t'}^{\text{pr}}, \mathbf{c}^{\text{pr}}, {t'}) \rVert ] ,
\end{equation}
where $\lambda$ is a weighting parameter, and $\mathbf{x}_{t}^{\text{pr}}$ and $\mathbf{c}^{\text{pr}}$ come from the prior dataset. DreamBooth is especially useful for personalized content generation where subject fidelity is critical.

{\bf Custom diffusion}~\citep{kumari2022customdiffusion} follows a similar approach but updates only the key and value projections in cross-attention layers, introducing a new token rather than reusing existing ones.
Prior preservation loss is maintained, since language drift is still experienced otherwise.

{\bf Customized Model Set} In our experiments, we will evaluate Dreambooth and Custom Diffusion. We adapt both these models to 10 different concepts based on 5 images per concept. The concepts are `lamp', `vase', `person2',`person3',`cat',`dog',`lighthouse',`waterfall',`bike' and `car' taken from CustomConcept101~\citep{kumari2022customdiffusion}.  We will refer to these ten models for both DreamBooth and Custom Diffusion as the \emph{Customized Model Set}. We have experimentally found that 10 concepts is enough for robust generalization. For a scale analysis, please see Appendix~\ref{sec:exp_scaling}.

\subsection{Drift Correction} \label{sec:drift_correction}

The two studied approaches, Dreambooth and Custom Diffusion, apply finetuning to adapt to the new data: they mainly focus on how good the learned model is on the target data, and do not study the possible detrimental effects for other classes. The Dreambooth method includes a method called \textit{prior regularization}, which by replaying general instances of the concept being learned (see Eq.~\ref{eq:db_with_prior_reg}), helps to prevent the model from overfitting to the new data and ensures that the representation of the superclass remains stable. This same mitigation strategy is also applied in custom diffusion~\citep{kumari2022customdiffusion}.

In this paper, we propose another regularization technique that can be applied during new concept learning. The method is remarkably simple and is motivated from continual learning literature. This field has proposed a variety of methods to counter forgetting during the learning of new concepts~\citep{de2021continual}. 
Regularization methods aim to regularize the learning of new concepts in such a way that it does not change weights which were found relevant for previous tasks. The field differentiates between parameter regularization 
methods, like EWC~\citep{kirkpatrick2017overcoming} which directly learn an importance weight for all the network parameters, or functional (or data) regularization, like Learning-without-Forgetting~\citep{li2017learning,pan2020continual} which regularizes the weights indirectly by imposing a penalty on changes between the (intermediate) outputs of a previous and current model. 

We propose to apply a functional regularization loss to the network during the training of new concepts. Our loss, called \emph{drift correction loss}, constrains the difference between the outputs of the pre-trained and fine-tuned models when the new concept is not present in the prompt. It has the following form:
\begin{equation}
    \mathbb{E}_{\mathbf{\epsilon}, \mathbf{x}, \mathbf{c}, t} [w_t \lVert \mathbf{\epsilon} - \mathbf{\epsilon}_\theta (\mathbf{x}_t, \mathbf{c}, t) \rVert + \lambda w_{t'} \lVert \mathbf{\epsilon}_{\theta^*} (\mathbf{x}_{t'}^{\text{pr}}, \mathbf{c}^{\text{pr}}, {t'}) - \mathbf{\epsilon}_\theta (\mathbf{x}_{t'}^{\text{pr}}, \mathbf{c}^{\text{pr}}, {t'}) \rVert ] ,\label{eq:fr}
\end{equation}
where the second term is the distillation loss, $\lambda$ is a relative weighting parameter and $\mathbf{\epsilon}_{\theta^*}$ is the base model.
This loss helps to maintain consistency in the model’s internal representations while allowing it to learn new information effectively. For the training process, we choose instances from the same class as the concept being learned, similar to those used by prior regularization. The change between our proposed drift correction method (Eq.~\ref{eq:fr}) and the existing prior regularization (Eq.~\ref{eq:db_with_prior_reg}) is that we do not require the finetuned network to estimate the true forward noise, but instead we want it to estimate the same noise as the original starting network. We will see that this small change significantly improves stability and mitigates forgetting.

In our evaluation, we provide results for \emph{DreamBooth (DB)} which includes the prior regularization, for \emph{Dreambooth with Drift Correction (DB-DC)} which also includes the prior regularization and for \emph{DreamBooth with Drift Correction without the prior regularization (DB-DC$\backslash$pr)}. Similarly, we show results for the various variants of Custom Diffusion (\emph{CD, CD-DC}, and \emph{CD-DC$\backslash$pr}). We also evaluate more recent customization methods such as SVDiff~\cite{han2023svdiff} and BOFT~\cite{liu2024boft}; detailed results for these methods can be found in Appendix~\ref{sec:additional_results}.

\section{Open-World Forgetting in Generative Model Adaptation}

In this section, we explore the effects of finetuning on foundational image generation models, particularly how even slight modifications can significantly impair the model's ability to retain previously acquired knowledge. We hypothesize that this degradation affects not only the model's performance on newly introduced tasks, but also its capacity to accurately reproduce or classify previously learned concepts. Given the broad scope of knowledge encompassed by the pretrained model, we refer to this phenomenon as \emph{open-world forgetting}.

\begin{figure}[tb]
    \centering
    \includegraphics[width=.8\linewidth]{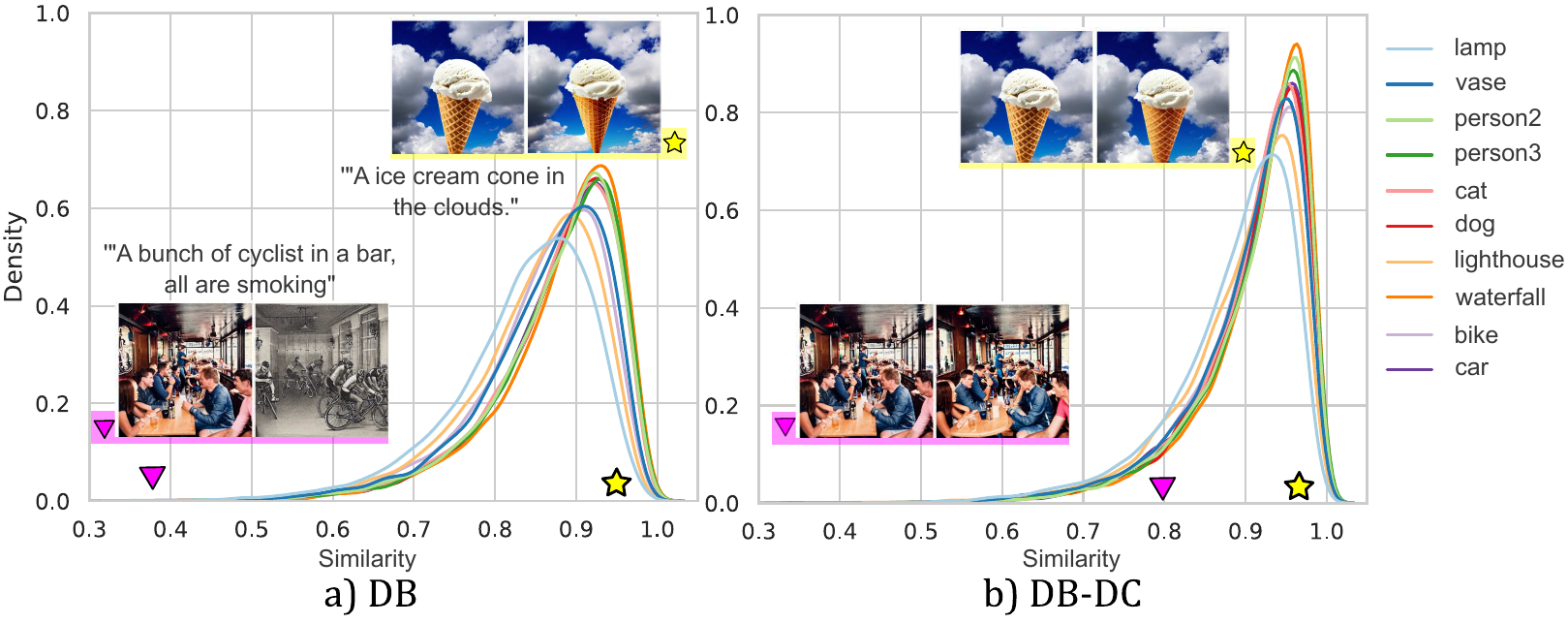}
    \caption{Similarity (measured as cosine distance in CLIP-I embedding space) between models before and after adaptation. Each curve represents one of the 10 models from the Customized Model Set. 
    a) Results with DreamBooth adaptation (includes prior regularization). b) Results with DreamBooth with Drift Correction. 
    For more results see Appendix~\ref{appx:more_samples}.}
    \vspace{-3mm}
    \label{fig:semantic_drift_general}
\end{figure}

As an initial experiment, to assess open-world forgetting, we evaluate both the original unaltered model (called \emph{base model} from now on) and the \emph{Customized Model Set} on 10,000 user prompts from DiffusionDB~\citep{wangDiffusionDBLargescalePrompt2023} dataset (prompt examples and further experimental details are provided in Appendix~\ref{appx:drift_exp_setup}).
Specifically, in Figure~\ref{fig:semantic_drift_general}, we measure the change of the resulting images using the cosine distance between CLIP-I encodings~\citep{radford2021clip} when generating images with the same prompt and seed. Distances in the CLIP-I embedding are related to semantic similarity between images, with smaller distances indicating more similar visual content and larger distances suggesting more significant differences in the generated images. A personalization method that does not alter the model would yield identical image outputs, resulting in a plot density concentrated at $1$. 

When considering Figure~\ref{fig:semantic_drift_general}a, we observe that even though most of the prompts from DiffusionDB are not related with the selected trained concepts, there is a significant part of the distribution that is shifted to the left. This shows indeed that the representations of the original model have changed. Furthermore, further analysis shows that open-world forgetting significantly alters the output in different ways, as illustrated by the samples in Figure~\ref{fig:semantic_drift_general}a. For instance, a sampled pair from the most dissimilar outputs (purple triangle) shows a complete change in content, colors, and scene composition that no longer matches the prompt. In contrast, a very similar pair (yellow star) closely adheres to the original model's output, with only changes in color or details\footnote{Similar curves are observed for different model families (see Appendix~\ref{sec:model_families}).}. Interestingly, when looking at Figure~\ref{fig:semantic_drift_general}b where we apply the proposed Drift Correction to DreamBooth, the distribution shifts to the right, showing that the drift has been reduced considerably. 

To better assess the impact of open-world forgetting, we propose to categorize the effects into two distinct types: \textbf{semantic} and \textbf{appearance drift}. 
Semantic drift implies a change at the class or object level, where one concept is effectively misencoded as another. Appearance drift, on the other hand, refers to shifts in the appearance of a concept that do not necessarily imply a change in recognition (e.g., alterations in color, texture, or scene composition). It is important to note that these two categories are highly correlated, and changes in either of them impact the other. 

\subsection{Semantic Drift} \label{sec:forgetting:semantic_drift}

Semantic drift represents fundamental changes in how the model understands and represents concepts, causing the generation of semantically divergent content following customization. In the experiment depicted in Figure~\ref{fig:semantic_drift_general}a, almost all prompts exhibit some level of drift, with a notable long tail of highly dissimilar generations. Many of these pronounced deviations have resulted in the generation of content that semantically no longer align with the input prompt. 

To evaluate how semantic drift affects generative models, we use a straightforward approach: we utilize the model's internal representations on different classification tasks \citep{pmlr-v202-mittal23a,tang2023emergent}. It is based on a recent insight that diffusion models can be directly applied for zero-class classification by leveraging the conditional likelihood estimation of the model. This provides a direct measure of how well the model maintains its understanding of various concepts. Concretely, we use Diffusion Classifier~\citep{li2023diffusion}, where a posterior distribution over classes $\{\mathbf{c}_i\}_{i=1}^N$ is calculated as:
\begin{equation}
    p_\theta(\mathbf{c}_i \mid  \mathbf{x})
    = \frac{\exp\{- \mathbb{E}_{t, \epsilon}[\|\epsilon - \epsilon_\theta(\mathbf{x}_t, \mathbf{c}_i)\|^2]\}}{\sum_{j=1}^{N} \exp\{- \mathbb{E}_{t, \epsilon}[\|\epsilon - \epsilon_\theta(\mathbf{x}_t, \mathbf{c}_j)\|^2]\}} .
\end{equation}
Monte Carlo sampling is performed over $t_i$ and $\epsilon$ to obtain a classifier from the model $\epsilon_\theta$.

While this method offers a simple, parameter-free approach to evaluating semantic drift, it is worth noting that alternative techniques have been proposed to assess the representation space of diffusion models. These include linear probing on activations \citep{ddae2023}, analysis of hierarchical features \citep{mukhopadhyay2023textfree}, and methods requiring a preliminary likelihood maximization stage \citep{chen2023robust}. However, these alternatives often involve additional computational steps or are subject to specific settings, potentially limiting their applicability or introducing complexity to the evaluation process.

We conducted zero-shot classification experiments across multiple datasets spanning diverse domains to quantify the semantic drift of the models. We perform two measurements. First, we measure the \emph{average zero-shot classification score} for the various models (the results are averaged over the 10 models of the Customized Model Set). Second, we establish the performance of the original pretrained model as the baseline, and measure the presence of semantic drift by calculating the drop in accuracy from the baseline. We also report the \emph{worst class drop} which is the drop in accuracy of the class that has suffered the largest deterioration due to the adaptation. 
For further details on the classification method, please refer to Appendix~\ref{appx:diffusion_classifier}. 

\begin{table}[t]
    \caption{\textbf{Average zero-shot classification} using the T2I models of the Customized Model Set for several image classification datasets, with concept and prompt fidelity metrics. Worst class drop between parenthesis. \textbf{Concept fidelity} (DINO, CLIP-I) \textbf{and prompt fidelity} (CLIP-T). Drift Correction maintains fidelity across metrics.}
    \label{tab:diffusion_classifier_and_concept_evaluation}
    \begin{center}
    \resizebox{\linewidth}{!}{
    \begin{tabular}{@{}lccccccc|ccc@{}}
    \toprule
    & CIFAR10 & STL10 & Flowers & Pets & ObjectNet & Food & Aircraft & DINO & CLIP-I & CLIP-T \\ \midrule
    Base Model & 81.60 & 93.00 & 50.00 & 86.87 & 28.50 & 71.09 & 23.40 & - & - & - \\
    \hdashline
    DB & 75.92 {\scriptsize (32.40)} & 91.30 {\scriptsize (18.60)} & 46.61 {\scriptsize (64.00)} & 82.61 {\scriptsize (36.43)} & 25.26 {\scriptsize (56.00)} & 65.48 {\scriptsize (56.00)} & 19.36 {\scriptsize (58.00)} & 0.424 & 0.676 & 0.790 \\
    DB-DC & 80.98 {\scriptsize (14.00)} & 93.36 {\scriptsize (4.40)} & 49.29 {\scriptsize (42.00)} & 86.64 {\scriptsize (17.14)} & 27.72 {\scriptsize (42.00)} & 69.07 {\scriptsize (44.00)} & 21.42 {\scriptsize (48.00)} & 0.428 & 0.682 & 0.780 \\
    DB-DC$\backslash$pr& 80.60 {\scriptsize (14.00)} & 92.94 {\scriptsize (5.20)} & 49.06 {\scriptsize (40.00)} & 86.37 {\scriptsize (16.43)} & 27.45 {\scriptsize (46.00)} & 68.79 {\scriptsize (44.00)} & 21.54 {\scriptsize (44.00)} & 0.432 & 0.684 & 0.778 \\
    \hdashline
    CD & 79.98 {\scriptsize (17.00)} & 91.40 {\scriptsize (12.20)} & 47.65 {\scriptsize (66.00)} & 83.46 {\scriptsize (33.57)} & 25.75 {\scriptsize (58.00)} & 65.25 {\scriptsize (56.00)} & 19.44 {\scriptsize (58.00)} & 0.442 & 0.693 & 0.791 \\
    CD-DC & 82.36 {\scriptsize (9.00)} & 93.02 {\scriptsize (5.00)} & 49.33 {\scriptsize (42.00)} & 86.37 {\scriptsize (16.43)} & 27.91 {\scriptsize (42.00)} & 69.19 {\scriptsize (44.00)} & 21.94 {\scriptsize (46.00)} & 0.438 & 0.693 & 0.790 \\
    CD-DC$\backslash$pr& 82.04 {\scriptsize (10.80)} & 92.76 {\scriptsize (6.00)} & 49.16 {\scriptsize (44.00)} & 86.70 {\scriptsize (20.00)} & 27.77 {\scriptsize (42.00)} & 68.99 {\scriptsize (44.00)} & 21.56 {\scriptsize (48.00)} & 0.438 & 0.693 & 0.787 \\
    \bottomrule
    \end{tabular}
    }
    \end{center}
\end{table}

The results in Table~\ref{tab:diffusion_classifier_and_concept_evaluation} are surprising, average zero-shot classification accuracy drop significantly on all the datasets: adapting a huge generative image foundation model to just five images of a new concept has a vast impact throughout the latent space of the diffusion models. When applying DreamBooth, average zero-shot performance drops by over 4\% on CIFAR10, Pets, Food and Aircraft. If we look at individual classes, the impact can be much larger. As indicated by the \emph{worst class drop}, for some classes, zero-shot performance drops by over 60\% (e.g. `vacuum cleaner' gets recognized as `microwave', `drill' or `laptop'). We show that these drops in performance are mitigated to a large extent by our alternative Drift Correction results (see DB-DC and CD-DC) and their average zero-shot classification scores are in general within 1\% of the base model. Removing the prior regularization from our method (see DB-DC$\backslash$pr and CD-DC$\backslash$pr) leads to only slightly lower results, showing the impact of our proposed regularization method. Also, worst class drop significantly reduces when applying DC, but for some datasets remains still high. 

Furthermore, to measure generation quality, diversity, and alignment between image and text of the learned concepts, we employ metrics adapted from  personalization methods: DINO, CLIP-I, and CLIP-T. In Table~\ref{tab:diffusion_classifier_and_concept_evaluation} we can see that the proposed regularization method DC does not negatively impact the image generation quality of the learned concepts.

Standard deviations and more zero-shot evaluation for more recent methods, including BOFT~\cite{liu2024boft} and SVDiff~\cite{han2023svdiff} are provided in Appendix~\ref{sec:additional_semantic}.

\subsection{Appearance Drift} \label{sec:forgetting:appearance_drift}

\begin{figure}[tb]
    \centering
    \includegraphics[width=0.9\linewidth]{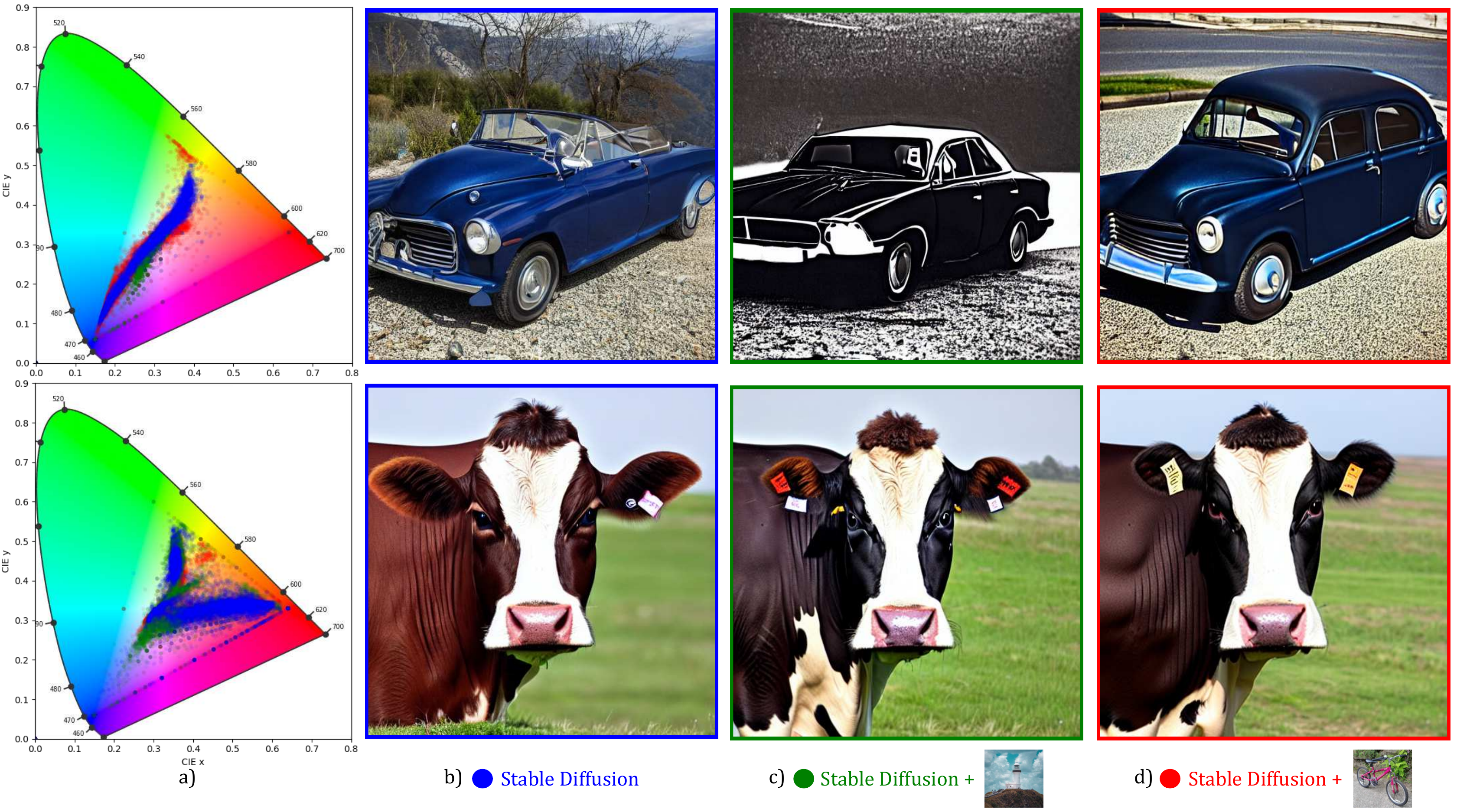}
    \caption{Appearance drift as consequence of DreamBooth customization. \textbf{a)} chromaticity plot of pixels of three realization of the prompts ('photo of a car/cow')  and the same seed with different models, namely \textbf{b)} the base model, \textbf{c)} model adapted to \textcolor{darkgreen}{lighthouse} and d) model adapted to \textcolor{red}{bike}.}
    \label{fig:drift_color}
\end{figure}

While open-world forgetting does not always result in significant changes to the core content of the image, as shown in Figure~\ref{fig:semantic_drift_general}a, it notably affects intra-class variation, color distribution, and texture characteristics. We define these collective changes as \textit{appearance drift}, a phenomenon that alters the model's representation space in subtle yet impactful ways.
Figure~\ref{fig:drift_color} demonstrates two key aspects of appearance drift; intra-class and contextual variation (first row), where different customizations of the base model lead to changes in car brand and background, while maintaining the overall concept of `car'. Color shift (second row), where the color palette of the generated images changes significantly, even when the intra-class characteristics and background remain relatively constant.

Appearance drift manifests through alterations in visual attributes at varying degrees of intensity. Finetuning can cause the model to reinterpret these visual features, leading to inconsistencies between original and newly generated outputs. Although initially subtle, appearance drift can substantially impact customized models. For example, when attempting to learn and generate a set of new concepts within the same context (e.g., for synthetic dataset creation or advertising purposes), each customization of the base model may result in color and content changes. This variability makes it challenging to achieve consistent results across multiple iterations.
Moreover, as the customization process alters the model's manifold, the resulting model becomes less reliable in domains outside the scope of the customization training images. This limitation highlights the importance of understanding and mitigating appearance drift in applications of fine-tuned text-to-image models. 

\paragraph{How to measure appearance drift?} Quantifying appearance drift presents unique challenges due to the inherent variability in text-to-image model outputs. Traditional metrics like LPIPS~\citep{zhang2018unreasonable} and DIST~\citep{ding2020image} are designed for image pair comparisons. However, the inherent variability in T2I model outputs means that images generated from the same prompt can vary significantly due to changes in seed, model weights, and prompt interpretation. Comparing just two images fails to capture the full range of possible outputs and does not adequately represent the model's capabilities or biases. Consequently, conclusions drawn from such limited comparisons may lack statistical significance.

To address this variability, the research community has employed metrics that measure distances between probability distributions (e.g., FID~\citep{heusel2017gans}, KID~\citep{binkowski2018demystifying}) of real-world observations and generated data\footnote{In general FID and KID require a set of real images for comparison. In our study, we consider the images generated by the original model as the ``real'' set, as we are measuring the shift from this initial distribution.}. In addition to these metrics, we also propose a new metric that directly measures the color drift between image sets.

\paragraph{Color Drift Index.} 
With the aim to introduce a clearly interpretable difference measure for the color content of the generated images, we propose a novel approach that measures specific properties in pixel space. Our method focuses on color assessment, as traditional image generation metrics tend to be more sensitive to intra-class and texture variations. Inspired by natural image statistics, we introduce the Color Drift Index (CDI), which estimates the color distribution associated with a specific concept by analyzing a large number of generated images. 

We utilize the CIE chromaticity diagram\footnote{Our approach offers the added benefit of being applicable across multiple color spaces.}, where each pixel color is projected onto a lobe-shaped space representing all visible colors.
Given a set of images $I = \{\mathbf{x}_i\}$ and their density distribution in the CIE chromaticity diagram $p^\text{CIE}( I )$, we calculate the CDI as the Wasserstein distance~\citep{panaretos2019statistical} between the color distributions of two sets of images:
\begin{equation}
   \text{CDI} (I_a, I_b) = W_p ( p^\text{CIE}( I_a ), p^\text{CIE}( I_b ) ) .
\label{eq:cdi}
\end{equation}

\begin{figure}[tb]
    \centering
    \includegraphics[width=0.9\linewidth]{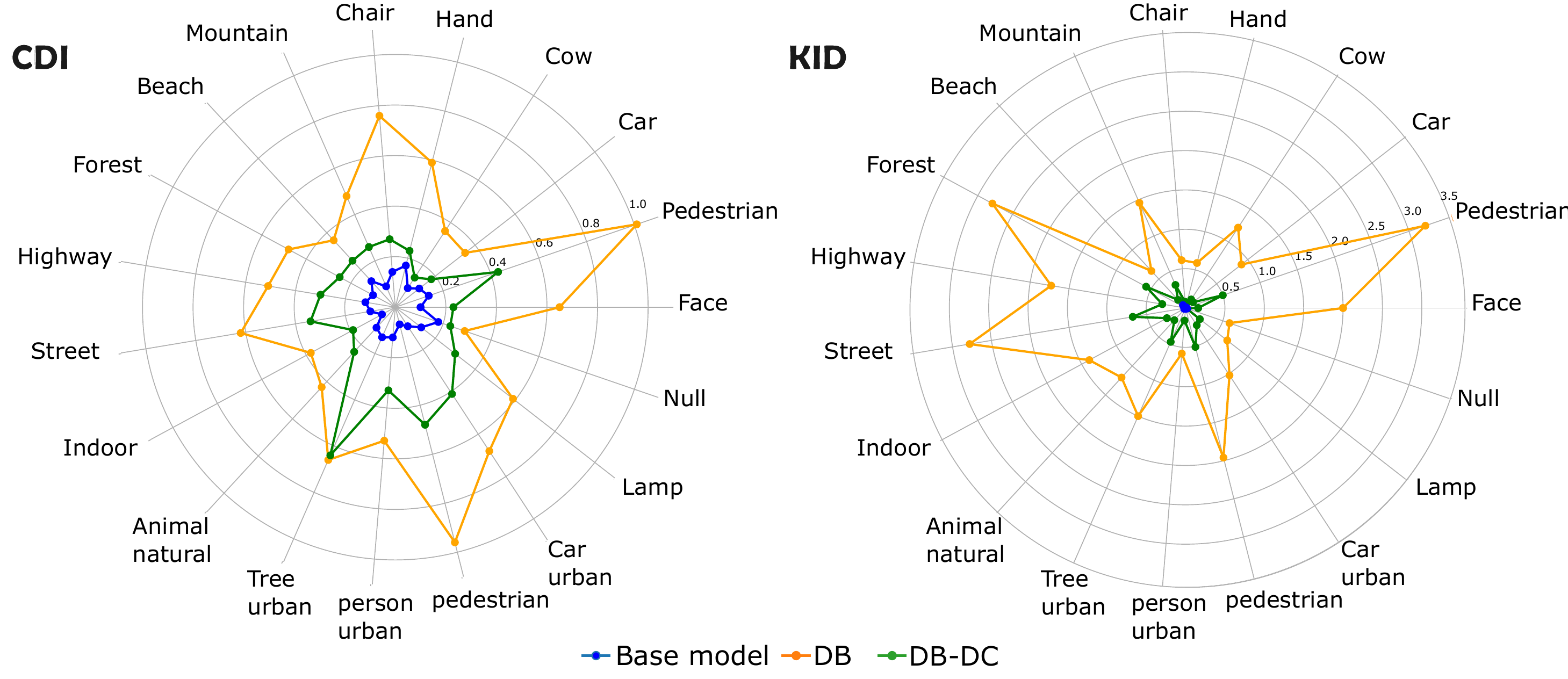}
    \caption{Appearance drift as consequence of customization measured with (left) Color Drift Index (CDI) and (right) Kernel Inception Distance (KID). The \textcolor{orange}{orange} and \textcolor{darkgreen}{green} line represent the distance between the base model and the customized model. The \textcolor{blue}{blue} line is a control line, representing the distance between two sets of images generated from different seeds both with the base model. Lines close to the origin are better. }
    
    \label{fig:cdi_radar}
\end{figure}

We evaluate the appearance drift using the CDI together with KID (FID results are presented in Appendix~\ref{appx:appearance_drift_full}).
We conducted a comprehensive experiment adopting the carefully curated selection of common concepts from the dataset of~\cite{torralba2003statistics}.
For each prompt, we generated 1,000 images using both original and the ten models from the Customized Model Set.
Figure~\ref{fig:cdi_radar} presents the mean values of the metrics across several adaptations, providing a visual representation of the differences captured by each measure. 
To help interpret the KID and CDI scores, we provide a control setting (blue line).  In this configuration, we measure CDI and KID between images generated with the base model but using different seeds (functioning as a lower bound). If we are sampling from the same distribution, the \textit{base model} should yield lower distances (approaching zero as the number of samples grows) than the customized models.

The results in Figure~\ref{fig:cdi_radar} reveal two key insights.
First, the \textit{base model} consistently produces smaller distance values compared to DB, confirming that the distribution of each concept is indeed changing due to appearance drift. Second, each concept is affected differently by the drift, attributable to the fact that each concept relates to different parts of the model's manifold. Furthermore, as demonstrated in Appendix~\ref{sec:ablations}, the magnitude and nature of the drift vary as a function of the content in the replay buffer and training images. Also, importantly, Figure~\ref{fig:cdi_radar} shows that our proposed method (DC) significantly reduces the impact of the appearance drift introduced by customization methods. Especially, the drift measure in KID is considerably reduced. 

For a detailed overview of the results, including individual model performances and additional customization methods, refer to Tables~\ref{tab:appereance_metric_comparison_simple} and \ref{tab:appereance_metric_comparison} in Appendix~\ref{appx:appearance_drift_full}.

These findings underscore the complexity and pervasiveness of appearance drift in fine-tuned text-to-image models. They highlight the need for robust mitigation strategies and careful consideration when deploying customized models in real-world applications, emphasizing the importance of ongoing research in this area to ensure the reliability and consistency of generated outputs.

\paragraph{User Study}
We conducted a user study to verify the effectiveness of the proposed drift correction to mitigate open-world forgetting. 
We evaluate over 100 image pairs (before/after customization) for both DreamBooth and Custom Diffusion with and without drift correction performed with 24 participants across 20 concepts for both DreamBooth and Custom Diffusion.
The results strongly favor our approach, with participants preferring our drift-corrected outputs 66\% of the time compared to baseline customization. 
The statistical analysis confirms the significance of these results (paired t-test: $t(23)=19.624$, $p < 0.001$; Wilcoxon signed-rank test: $p < 0.001$). 
We have included the settings and sample images of the user study in Appendix~\ref{sec:user_study}. 

\subsection{Local Drift} \label{sec:forgetting:local_drift}
In this paper, we have focused on drift throughout the whole diffusion model manifold. Previous works, especially those in the machine unlearning community~\citep{gandikota2023erasing}, have concentrated on \emph{local drift}. When removing a concept from a model, it is believed to mainly impact the representation of closely related concepts (hence the name local drift).
Our findings suggest that the effects of finetuning are more pervasive than previously thought, potentially influencing the model's understanding and representation of far-away categories as well as close-by (local) categories. 

\begin{figure}[ht]
    \centering
    \includegraphics[width=1\linewidth]{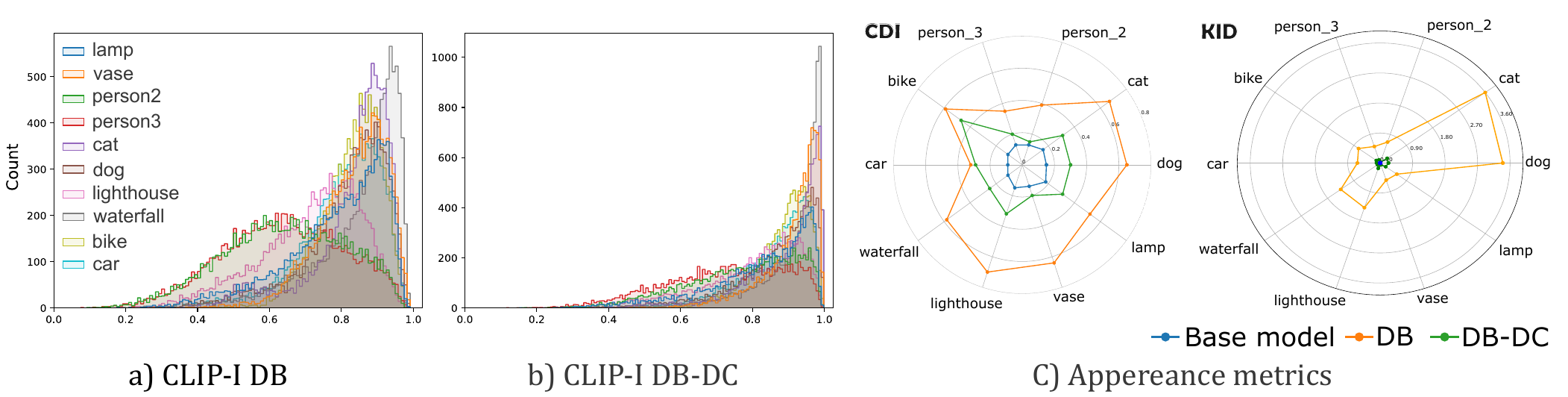}
    \caption{
    Similarity (measured as cosine distance in CLIP-I embedding space) and perceptual metrics between models before and after adaptation. For each concept trained, we evaluate closely related concepts to measure the local drift.
    \textbf{a)} Results with DreamBooth adaptation (includes prior regularization). \textbf{b)} Results with DreamBooth with Drift Correction. \textbf{c)} Color Drift Index (CDI) and Kernel Inception Distance (KID).
    For more results see Appendix~\ref{appx:more_samples}.
    }    
    \label{fig:semantic_drift_superclass_general}
\end{figure}

Here, we repeated our experiments from Section~\ref{sec:forgetting:semantic_drift} and \ref{sec:forgetting:appearance_drift} to measure the local semantic and local appearance drift. For this setup, we generated 1,000 samples of the closest concepts (superclasses) to each trained model (see Appendix~\ref{appx:drift_exp_setup} for the details) and evaluated the CLIP-I, CDI, FID, and KID metrics. As Figure~\ref{fig:semantic_drift_superclass_general}a shows, the semantic drift is showing a significant shift towards the left, indicating that local drift is more pronounced. For appearance drift, Figure~\ref{fig:semantic_drift_superclass_general}c depicts a more uniform color and KID shift over all the models; this shows that related concepts are affected with a similar magnitude by appearance drift. Again the application of our proposed Drift Correction method greatly reduces both the local semantic drift (as measured in Figure~\ref{fig:semantic_drift_superclass_general}b and it almost removes the local appearance drift as measured by KID, even though some color drift remains (see Figure~\ref{fig:semantic_drift_superclass_general}c).

\section{Discussion and Conclusion}

Our investigation into unintended consequences of generative model adaptation reveals several key findings. First, we demonstrate that finetuning foundational generative models leads to substantial \textit{open-world forgetting}, manifesting as both \textit{semantic} and \textit{appearance drift}. Our results show that even minor adaptations can cause significant deterioration in the model's ability to maintain its original capabilities.
To quantify these effects, we introduced novel evaluation approaches: measuring semantic drift through zero-shot classification performance across diverse image classification tasks, and assessing appearance drift through our proposed Color Drift Index combined with traditional metrics like KID. These methods provide a framework for understanding and measuring the impact of model adaptation on both semantic understanding and visual representation.
Additionally, we propose a technique to mitigate open-world forgetting using functional regularization. Our experiments demonstrate that this method effectively preserves foundational knowledge while allowing for successful customization, offering a promising direction for developing more robust adaptation techniques. 

We have focussed our analysis on the forgetting of customization methods, however, the adaptation of diffusion models occurs in various domains. The understanding and addressing of open-world forgetting is also important in these fields, and we believe 
the developed metrics can be useful. As a preliminary analysis, we have included results in Appendix~\ref{sec:areas_forgetting} for \emph{unlearning}, where concepts are removed from a diffusion model, and \emph{quantization}, where quantization is applied to obtain a more compute-efficient version of the model. The initial results confirm that indeed these operations lead to forgetting, and our metrics can be used to study this phenomenon.
Another future research direction which we aim to explore is the development of optimization methods to identify the most affected areas in the latent space during adaptation. 

\newpage
\section*{Ethical Statement}
We acknowledge the potential ethical implications of deploying generative models, including issues related to privacy, data misuse, and the propagation of biases. All models used in this paper are publicly available, as well as the base training scripts. We will release the modified codes to reproduce the results of this paper. We also want to point out the potential role of customization approaches in the generation of fake news, and we encourage and support responsible usage of these models. Finally, we think that awareness of open-world forgetting can contribute to safer models in the future, since it encourages a more thorough investigation into the unpredictable changes occurring when adapting models to new data. 

\section*{Reproducibility Statement}
To facilitate reproducibility, we will make the entire source code and scripts needed to replicate all results presented in this paper available after the peer review period. We will release the code for the novel color metric we have introduced. We conducted all experiments using publicly accessible datasets. Elaborate details of all experiments have been provided in the Appendices. 

\bibliography{longstrings,iclr2025_conference}

\begin{thebibliography}{54}
\providecommand{\natexlab}[1]{#1}
\providecommand{\url}[1]{\texttt{#1}}
\expandafter\ifx\csname urlstyle\endcsname\relax
  \providecommand{\doi}[1]{doi: #1}\else
  \providecommand{\doi}{doi: \begingroup \urlstyle{rm}\Url}\fi

\bibitem[Bi{\'n}kowski et~al.(2018)Bi{\'n}kowski, Sutherland, Arbel, and Gretton]{binkowski2018demystifying}
M~Bi{\'n}kowski, DJ~Sutherland, M~Arbel, and A~Gretton.
\newblock Demystifying mmd gans.
\newblock In \emph{ICLR}, 2018.

\bibitem[Brack et~al.(2023)Brack, Friedrich, Hintersdorf, Struppek, Schramowski, and Kersting]{brack2023sega}
Manuel Brack, Felix Friedrich, Dominik Hintersdorf, Lukas Struppek, Patrick Schramowski, and Kristian Kersting.
\newblock Sega: Instructing text-to-image models using semantic guidance.
\newblock \emph{Advances in Neural Information Processing Systems}, 36:\penalty0 25365--25389, 2023.

\bibitem[Caron et~al.(2021)Caron, Touvron, Misra, J{\'e}gou, Mairal, Bojanowski, and Joulin]{caron2021emerging}
Mathilde Caron, Hugo Touvron, Ishan Misra, Herv{\'e} J{\'e}gou, Julien Mairal, Piotr Bojanowski, and Armand Joulin.
\newblock Emerging properties in self-supervised vision transformers.
\newblock In \emph{Proceedings of the IEEE/CVF International Conference on Computer Vision}, pp.\  9650--9660, 2021.

\bibitem[Chen et~al.(2023)Chen, Dong, Wang, Yang, Duan, Su, and Zhu]{chen2023robust}
Huanran Chen, Yinpeng Dong, Zhengyi Wang, Xiao Yang, Chengqi Duan, Hang Su, and Jun Zhu.
\newblock Robust classification via a single diffusion model.
\newblock \emph{arXiv preprint arXiv:2305.15241}, 2023.

\bibitem[Cossu et~al.(2024)Cossu, Carta, Passaro, Lomonaco, Tuytelaars, and Bacciu]{cossu2024continual}
Andrea Cossu, Antonio Carta, Lucia Passaro, Vincenzo Lomonaco, Tinne Tuytelaars, and Davide Bacciu.
\newblock Continual pre-training mitigates forgetting in language and vision.
\newblock \emph{Neural Networks}, 179:\penalty0 106492, 2024.

\bibitem[Daras \& Dimakis(2022)Daras and Dimakis]{daras2022multiresolution}
Giannis Daras and Alex Dimakis.
\newblock Multiresolution textual inversion.
\newblock In \emph{NeurIPS 2022 Workshop on Score-Based Methods}, 2022.

\bibitem[De~Lange et~al.(2021)De~Lange, Aljundi, Masana, Parisot, Jia, Leonardis, Slabaugh, and Tuytelaars]{de2021continual}
Matthias De~Lange, Rahaf Aljundi, Marc Masana, Sarah Parisot, Xu~Jia, Ale{\v{s}} Leonardis, Gregory Slabaugh, and Tinne Tuytelaars.
\newblock A continual learning survey: Defying forgetting in classification tasks.
\newblock \emph{IEEE transactions on pattern analysis and machine intelligence}, 44\penalty0 (7):\penalty0 3366--3385, 2021.

\bibitem[Dettmers et~al.(2022)Dettmers, Lewis, Belkada, and Zettlemoyer]{dettmers2022gpt3}
Tim Dettmers, Mike Lewis, Younes Belkada, and Luke Zettlemoyer.
\newblock Llm.int8(): 8-bit matrix multiplication for transformers at scale.
\newblock \emph{Advances in Neural Information Processing Systems}, 35:\penalty0 30318--30332, 2022.

\bibitem[Dettmers et~al.(2024)Dettmers, Pagnoni, Holtzman, and Zettlemoyer]{dettmers2024qlora}
Tim Dettmers, Artidoro Pagnoni, Ari Holtzman, and Luke Zettlemoyer.
\newblock Qlora: Efficient finetuning of quantized llms.
\newblock \emph{Advances in Neural Information Processing Systems}, 36, 2024.

\bibitem[Ding et~al.(2020)Ding, Ma, Wang, and Simoncelli]{ding2020image}
Keyan Ding, Kede Ma, Shiqi Wang, and Eero~P Simoncelli.
\newblock Image quality assessment: Unifying structure and texture similarity.
\newblock \emph{IEEE transactions on pattern analysis and machine intelligence}, 44\penalty0 (5):\penalty0 2567--2581, 2020.

\bibitem[Dong et~al.(2022)Dong, Wei, and Lin]{dong2022dreamartist}
Ziyi Dong, Pengxu Wei, and Liang Lin.
\newblock Dreamartist: Towards controllable one-shot text-to-image generation via contrastive prompt-tuning.
\newblock \emph{arXiv preprint arXiv:2211.11337}, 2022.

\bibitem[Esser et~al.(2024)Esser, Kulal, Blattmann, Entezari, M{\"u}ller, Saini, Levi, Lorenz, Sauer, Boesel, et~al.]{esser2024scaling}
Patrick Esser, Sumith Kulal, Andreas Blattmann, Rahim Entezari, Jonas M{\"u}ller, Harry Saini, Yam Levi, Dominik Lorenz, Axel Sauer, Frederic Boesel, et~al.
\newblock Scaling rectified flow transformers for high-resolution image synthesis.
\newblock In \emph{Forty-first International Conference on Machine Learning}, 2024.

\bibitem[Gal et~al.(2023)Gal, Alaluf, Atzmon, Patashnik, Bermano, Chechik, and Cohen-Or]{textual_inversion}
Rinon Gal, Yuval Alaluf, Yuval Atzmon, Or~Patashnik, Amit~H Bermano, Gal Chechik, and Daniel Cohen-Or.
\newblock An image is worth one word: Personalizing text-to-image generation using textual inversion.
\newblock \emph{ICLR}, 2023.

\bibitem[Gandikota et~al.(2023)Gandikota, Materzynska, Fiotto-Kaufman, and Bau]{gandikota2023erasing}
Rohit Gandikota, Joanna Materzynska, Jaden Fiotto-Kaufman, and David Bau.
\newblock Erasing concepts from diffusion models.
\newblock In \emph{Proceedings of the IEEE/CVF International Conference on Computer Vision}, pp.\  2426--2436, 2023.

\bibitem[Gong et~al.(2025)Gong, Chen, Wei, Chen, and Jiang]{gong2025reliable}
Chao Gong, Kai Chen, Zhipeng Wei, Jingjing Chen, and Yu-Gang Jiang.
\newblock Reliable and efficient concept erasure of text-to-image diffusion models.
\newblock In \emph{European Conference on Computer Vision}, pp.\  73--88. Springer, 2025.

\bibitem[Han et~al.(2023{\natexlab{a}})Han, Yang, Kwon, and Ye]{hiper2023}
Inhwa Han, Serin Yang, Taesung Kwon, and Jong~Chul Ye.
\newblock Highly personalized text embedding for image manipulation by stable diffusion.
\newblock \emph{arXiv preprint arXiv:2303.08767}, 2023{\natexlab{a}}.

\bibitem[Han et~al.(2023{\natexlab{b}})Han, Li, Zhang, Milanfar, Metaxas, and Yang]{han2023svdiff}
Ligong Han, Yinxiao Li, Han Zhang, Peyman Milanfar, Dimitris Metaxas, and Feng Yang.
\newblock Svdiff: Compact parameter space for diffusion fine-tuning.
\newblock \emph{ICCV}, 2023{\natexlab{b}}.

\bibitem[Heusel et~al.(2017)Heusel, Ramsauer, Unterthiner, Nessler, and Hochreiter]{heusel2017gans}
Martin Heusel, Hubert Ramsauer, Thomas Unterthiner, Bernhard Nessler, and Sepp Hochreiter.
\newblock Gans trained by a two time-scale update rule converge to a local nash equilibrium.
\newblock In \emph{NeurIPS}, pp.\  6626--6637, 2017.

\bibitem[Hu et~al.(2022)Hu, Shen, Wallis, Allen-Zhu, Li, Wang, Wang, and Chen]{hu2022lora}
Edward~J Hu, Yelong Shen, Phillip Wallis, Zeyuan Allen-Zhu, Yuanzhi Li, Shean Wang, Lu~Wang, and Weizhu Chen.
\newblock Lo{RA}: Low-rank adaptation of large language models.
\newblock In \emph{International Conference on Learning Representations}, 2022.
\newblock URL \url{https://openreview.net/forum?id=nZeVKeeFYf9}.

\bibitem[Kirkpatrick et~al.(2017)Kirkpatrick, Pascanu, Rabinowitz, Veness, Desjardins, Rusu, Milan, Quan, Ramalho, Grabska-Barwinska, et~al.]{kirkpatrick2017overcoming}
James Kirkpatrick, Razvan Pascanu, Neil Rabinowitz, Joel Veness, Guillaume Desjardins, Andrei~A Rusu, Kieran Milan, John Quan, Tiago Ramalho, Agnieszka Grabska-Barwinska, et~al.
\newblock Overcoming catastrophic forgetting in neural networks.
\newblock \emph{Proceedings of the national academy of sciences}, 114\penalty0 (13):\penalty0 3521--3526, 2017.

\bibitem[Kumari et~al.(2023{\natexlab{a}})Kumari, Zhang, Wang, Shechtman, Zhang, and Zhu]{kumari2023ablating}
Nupur Kumari, Bingliang Zhang, Sheng-Yu Wang, Eli Shechtman, Richard Zhang, and Jun-Yan Zhu.
\newblock Ablating concepts in text-to-image diffusion models.
\newblock \emph{arXiv preprint arXiv:2303.13516}, 2023{\natexlab{a}}.

\bibitem[Kumari et~al.(2023{\natexlab{b}})Kumari, Zhang, Zhang, Shechtman, and Zhu]{kumari2022customdiffusion}
Nupur Kumari, Bingliang Zhang, Richard Zhang, Eli Shechtman, and Jun-Yan Zhu.
\newblock Multi-concept customization of text-to-image diffusion.
\newblock \emph{CVPR}, 2023{\natexlab{b}}.

\bibitem[Labs(2023)]{flux2023}
Black~Forest Labs.
\newblock Flux.
\newblock \url{https://github.com/black-forest-labs/flux}, 2023.

\bibitem[Li et~al.(2023)Li, Prabhudesai, Duggal, Brown, and Pathak]{li2023diffusion}
Alexander~C. Li, Mihir Prabhudesai, Shivam Duggal, Ellis Brown, and Deepak Pathak.
\newblock Your diffusion model is secretly a zero-shot classifier.
\newblock In \emph{Proceedings of the IEEE/CVF International Conference on Computer Vision (ICCV)}, pp.\  2206--2217, October 2023.

\bibitem[Li et~al.(2025)Li, Lin, Zhang, Cai, Li, Guo, Xie, Meng, Zhu, and Han]{li2024svdquant}
Muyang Li, Yujun Lin, Zhekai Zhang, Tianle Cai, Xiuyu Li, Junxian Guo, Enze Xie, Chenlin Meng, Jun-Yan Zhu, and Song Han.
\newblock Svdquant: Absorbing outliers by low-rank components for 4-bit diffusion models.
\newblock In \emph{The Thirteenth International Conference on Learning Representations}, 2025.

\bibitem[Li \& Hoiem(2017)Li and Hoiem]{li2017learning}
Zhizhong Li and Derek Hoiem.
\newblock Learning without forgetting.
\newblock \emph{IEEE transactions on pattern analysis and machine intelligence}, 40\penalty0 (12):\penalty0 2935--2947, 2017.

\bibitem[Liu et~al.(2024)Liu, Qiu, Feng, Xiu, Xue, Yu, Feng, Liu, Heo, Peng, Wen, Black, Weller, and Sch{\"o}lkopf]{liu2024boft}
Weiyang Liu, Zeju Qiu, Yao Feng, Yuliang Xiu, Yuxuan Xue, Longhui Yu, Haiwen Feng, Zhen Liu, Juyeon Heo, Songyou Peng, Yandong Wen, Michael~J. Black, Adrian Weller, and Bernhard Sch{\"o}lkopf.
\newblock Parameter-efficient orthogonal finetuning via butterfly factorization.
\newblock In \emph{ICLR}, 2024.

\bibitem[Lopez-Paz \& Ranzato(2017)Lopez-Paz and Ranzato]{lopez2017gradient}
David Lopez-Paz and Marc'Aurelio Ranzato.
\newblock Gradient episodic memory for continual learning.
\newblock \emph{Advances in neural information processing systems}, 30, 2017.

\bibitem[Masana et~al.(2022)Masana, Liu, Twardowski, Menta, Bagdanov, and Van De~Weijer]{masana2022class}
Marc Masana, Xialei Liu, Bart{\l}omiej Twardowski, Mikel Menta, Andrew~D Bagdanov, and Joost Van De~Weijer.
\newblock Class-incremental learning: survey and performance evaluation on image classification.
\newblock \emph{IEEE Transactions on Pattern Analysis and Machine Intelligence}, 45\penalty0 (5):\penalty0 5513--5533, 2022.

\bibitem[Mittal et~al.(2023)Mittal, Abstreiter, Bauer, Sch\"{o}lkopf, and Mehrjou]{pmlr-v202-mittal23a}
Sarthak Mittal, Korbinian Abstreiter, Stefan Bauer, Bernhard Sch\"{o}lkopf, and Arash Mehrjou.
\newblock Diffusion based representation learning.
\newblock In Andreas Krause, Emma Brunskill, Kyunghyun Cho, Barbara Engelhardt, Sivan Sabato, and Jonathan Scarlett (eds.), \emph{Proceedings of the 40th International Conference on Machine Learning}, volume 202 of \emph{Proceedings of Machine Learning Research}, pp.\  24963--24982. PMLR, 23--29 Jul 2023.
\newblock URL \url{https://proceedings.mlr.press/v202/mittal23a.html}.

\bibitem[Mukhopadhyay et~al.(2023)Mukhopadhyay, Gwilliam, Yamaguchi, Agarwal, Padmanabhan, Swaminathan, Zhou, and Shrivastava]{mukhopadhyay2023textfree}
Soumik Mukhopadhyay, Matthew Gwilliam, Yosuke Yamaguchi, Vatsal Agarwal, Namitha Padmanabhan, Archana Swaminathan, Tianyi Zhou, and Abhinav Shrivastava.
\newblock Do text-free diffusion models learn discriminative visual representations?, 2023.

\bibitem[Nair et~al.(2023)Nair, Cherian, Lohit, Wang, Koike-Akino, Patel, and Marks]{nair2023steered}
Nithin~Gopalakrishnan Nair, Anoop Cherian, Suhas Lohit, Ye~Wang, Toshiaki Koike-Akino, Vishal~M Patel, and Tim~K Marks.
\newblock Steered diffusion: A generalized framework for plug-and-play conditional image synthesis.
\newblock In \emph{Proceedings of the IEEE/CVF International Conference on Computer Vision}, pp.\  20850--20860, 2023.

\bibitem[Pan et~al.(2020)Pan, Swaroop, Immer, Eschenhagen, Turner, and Khan]{pan2020continual}
Pingbo Pan, Siddharth Swaroop, Alexander Immer, Runa Eschenhagen, Richard Turner, and Mohammad Emtiyaz~E Khan.
\newblock Continual deep learning by functional regularisation of memorable past.
\newblock \emph{Advances in neural information processing systems}, 33:\penalty0 4453--4464, 2020.

\bibitem[Panaretos \& Zemel(2019)Panaretos and Zemel]{panaretos2019statistical}
Victor~M Panaretos and Yoav Zemel.
\newblock Statistical aspects of wasserstein distances.
\newblock \emph{Annual review of statistics and its application}, 6\penalty0 (1):\penalty0 405--431, 2019.

\bibitem[Radford et~al.(2021)Radford, Kim, Hallacy, Ramesh, Goh, Agarwal, Sastry, Askell, Mishkin, Clark, et~al.]{radford2021clip}
Alec Radford, Jong~Wook Kim, Chris Hallacy, Aditya Ramesh, Gabriel Goh, Sandhini Agarwal, Girish Sastry, Amanda Askell, Pamela Mishkin, Jack Clark, et~al.
\newblock Learning transferable visual models from natural language supervision.
\newblock In \emph{International conference on machine learning}, pp.\  8748--8763. PMLR, 2021.

\bibitem[Ramesh et~al.(2022)Ramesh, Dhariwal, Nichol, Chu, and Chen]{ramesh2022dalle2}
Aditya Ramesh, Prafulla Dhariwal, Alex Nichol, Casey Chu, and Mark Chen.
\newblock Hierarchical text-conditional image generation with clip latents.
\newblock \emph{arXiv preprint arXiv:2204.06125}, 2022.

\bibitem[Rombach et~al.(2022)Rombach, Blattmann, Lorenz, Esser, and Ommer]{Rombach_2022_CVPR_stablediffusion}
Robin Rombach, Andreas Blattmann, Dominik Lorenz, Patrick Esser, and Bj\"orn Ommer.
\newblock High-resolution image synthesis with latent diffusion models.
\newblock In \emph{Proceedings of the IEEE/CVF Conference on Computer Vision and Pattern Recognition (CVPR)}, pp.\  10684--10695, 06 2022.

\bibitem[Ruiz et~al.(2023)Ruiz, Li, Jampani, Pritch, Rubinstein, and Aberman]{ruiz2023dreambooth}
Nataniel Ruiz, Yuanzhen Li, Varun Jampani, Yael Pritch, Michael Rubinstein, and Kfir Aberman.
\newblock Dreambooth: Fine tuning text-to-image diffusion models for subject-driven generation.
\newblock In \emph{Proceedings of the IEEE/CVF Conference on Computer Vision and Pattern Recognition}, 2023.

\bibitem[Ruiz et~al.(2024)Ruiz, Li, Jampani, Wei, Hou, Pritch, Wadhwa, Rubinstein, and Aberman]{ruiz2024hyperdreambooth}
Nataniel Ruiz, Yuanzhen Li, Varun Jampani, Wei Wei, Tingbo Hou, Yael Pritch, Neal Wadhwa, Michael Rubinstein, and Kfir Aberman.
\newblock Hyperdreambooth: Hypernetworks for fast personalization of text-to-image models.
\newblock In \emph{Proceedings of the IEEE/CVF Conference on Computer Vision and Pattern Recognition}, pp.\  6527--6536, 2024.

\bibitem[Scialom et~al.(2022)Scialom, Chakrabarty, and Muresan]{scialom2022fine}
Thomas Scialom, Tuhin Chakrabarty, and Smaranda Muresan.
\newblock Fine-tuned language models are continual learners.
\newblock \emph{arXiv preprint arXiv:2205.12393}, 2022.

\bibitem[Shonenkov et~al.(2023)Shonenkov, Konstantinov, Bakshandaeva, Schuhmann, Ivanova, and Klokova]{deepfloyd}
Alex Shonenkov, Misha Konstantinov, Daria Bakshandaeva, Christoph Schuhmann, Ksenia Ivanova, and Nadiia Klokova.
\newblock Deepfloyd-if.
\newblock \url{https://github.com/deep-floyd/IF}, 2023.

\bibitem[Smith et~al.(2023)Smith, Hsu, Zhang, Hua, Kira, Shen, and Jin]{smith2023clora}
James~Seale Smith, Yen-Chang Hsu, Lingyu Zhang, Ting Hua, Zsolt Kira, Yilin Shen, and Hongxia Jin.
\newblock Continual diffusion: Continual customization of text-to-image diffusion with c-lora.
\newblock \emph{arXiv preprint arXiv:2304.06027}, 2023.

\bibitem[Sun et~al.(2024)Sun, Liang, Dong, Li, Ding, and Cong]{sun2024create_your_world}
Gan Sun, Wenqi Liang, Jiahua Dong, Jun Li, Zhengming Ding, and Yang Cong.
\newblock Create your world: Lifelong text-to-image diffusion.
\newblock \emph{IEEE Transactions on Pattern Analysis and Machine Intelligence}, 2024.

\bibitem[Tang et~al.(2023)Tang, Jia, Wang, Phoo, and Hariharan]{tang2023emergent}
Luming Tang, Menglin Jia, Qianqian Wang, Cheng~Perng Phoo, and Bharath Hariharan.
\newblock Emergent correspondence from image diffusion.
\newblock In \emph{Thirty-seventh Conference on Neural Information Processing Systems}, 2023.
\newblock URL \url{https://openreview.net/forum?id=ypOiXjdfnU}.

\bibitem[Torralba \& Oliva(2003)Torralba and Oliva]{torralba2003statistics}
Antonio Torralba and Aude Oliva.
\newblock Statistics of natural image categories.
\newblock \emph{Network: computation in neural systems}, 14\penalty0 (3):\penalty0 391, 2003.

\bibitem[von Platen et~al.(2022)von Platen, Patil, Lozhkov, Cuenca, Lambert, Rasul, Davaadorj, Nair, Paul, Berman, Xu, Liu, and Wolf]{von-platen-etal-2022-diffusers}
Patrick von Platen, Suraj Patil, Anton Lozhkov, Pedro Cuenca, Nathan Lambert, Kashif Rasul, Mishig Davaadorj, Dhruv Nair, Sayak Paul, William Berman, Yiyi Xu, Steven Liu, and Thomas Wolf.
\newblock Diffusers: State-of-the-art diffusion models.
\newblock \url{https://github.com/huggingface/diffusers}, 2022.

\bibitem[Voynov et~al.(2023)Voynov, Chu, Cohen-Or, and Aberman]{voynov2023ETI}
Andrey Voynov, Qinghao Chu, Daniel Cohen-Or, and Kfir Aberman.
\newblock $p+$: Extended textual conditioning in text-to-image generation.
\newblock \emph{arXiv preprint arXiv:2303.09522}, 2023.

\bibitem[Wang et~al.(2023)Wang, Montoya, Munechika, Yang, Hoover, and Chau]{wangDiffusionDBLargescalePrompt2023}
Zijie~J. Wang, Evan Montoya, David Munechika, Haoyang Yang, Benjamin Hoover, and Duen~Horng Chau.
\newblock {{DiffusionDB}}: {{A}} large-scale prompt gallery dataset for text-to-image generative models.
\newblock In \emph{Proceedings of the 61st Annual Meeting of the Association for Computational Linguistics (Volume 1: {{Long}} Papers)}, 2023.
\newblock URL \url{https://aclanthology.org/2023.acl-long.51}.

\bibitem[Xiang et~al.(2023)Xiang, Yang, Huang, and Wang]{ddae2023}
Weilai Xiang, Hongyu Yang, Di~Huang, and Yunhong Wang.
\newblock Denoising diffusion autoencoders are unified self-supervised learners.
\newblock In \emph{Proceedings of the IEEE/CVF International Conference on Computer Vision}, 2023.

\bibitem[Xie et~al.(2024)Xie, Chen, Chen, Cai, Tang, Lin, Zhang, Li, Zhu, Lu, and Han]{xie2024sana}
Enze Xie, Junsong Chen, Junyu Chen, Han Cai, Haotian Tang, Yujun Lin, Zhekai Zhang, Muyang Li, Ligeng Zhu, Yao Lu, and Song Han.
\newblock Sana: Efficient high-resolution image synthesis with linear diffusion transformer, 2024.
\newblock URL \url{https://arxiv.org/abs/2410.10629}.

\bibitem[Ye et~al.(2023)Ye, Zhang, Liu, Han, and Yang]{ye2023ip-adapter}
Hu~Ye, Jun Zhang, Sibo Liu, Xiao Han, and Wei Yang.
\newblock Ip-adapter: Text compatible image prompt adapter for text-to-image diffusion models.
\newblock \emph{arXiv preprint arXiv:2308.06721}, 2023.

\bibitem[Zhang \& Agrawala(2023)Zhang and Agrawala]{zhang2023controlnet}
Lvmin Zhang and Maneesh Agrawala.
\newblock Adding conditional control to text-to-image diffusion models.
\newblock \emph{arXiv preprint arXiv:2302.05543}, 2023.

\bibitem[Zhang et~al.(2018)Zhang, Isola, Efros, Shechtman, and Wang]{zhang2018unreasonable}
Richard Zhang, Phillip Isola, Alexei~A Efros, Eli Shechtman, and Oliver Wang.
\newblock The unreasonable effectiveness of deep features as a perceptual metric.
\newblock In \emph{Proceedings of the IEEE conference on computer vision and pattern recognition}, pp.\  586--595, 2018.

\bibitem[Zhou et~al.(2024)Zhou, Zhang, Gu, and Sun]{zhou2024customization}
Yufan Zhou, Ruiyi Zhang, Jiuxiang Gu, and Tong Sun.
\newblock Customization assistant for text-to-image generation.
\newblock In \emph{Proceedings of the IEEE/CVF Conference on Computer Vision and Pattern Recognition}, pp.\  9182--9191, 2024.

\end{thebibliography}
\bibliographystyle{iclr2025_conference}

%%%%%%%%%%%%%%%%%%%%%%%%%%%%%%% APPENDIX %%%%%%%%%%%%%%%%%%
\newpage
\appendix

\section{Additional Experiments} \label{sec:appx_exps}

\subsection{Analysis of Open-World Forgetting Across Model Modifications} \label{sec:areas_forgetting}

To demonstrate that open-world forgetting is a general phenomenon that occurs across different types of model modifications, we analyzed three broad categories of model alterations: model customization, unlearning, and quantization methods.

For customization, we evaluated DreamBooth~\cite{ruiz2023dreambooth}, Custom Diffusion~\cite{kumari2022customdiffusion}, and BOFT~\cite{liu2024boft}, which are commonly used for personalizing generative models, learning \textit{decoritems\_lamp1} concept. For unlearning, we examined NSFW and nudity content removal approaches including Erasing Concepts~\cite{gandikota2023erasing} and Ablating Concepts~\cite{kumari2023ablating}, as well as RECE~\cite{gong2025reliable}. Finally, for quantization, we compared LLM.int8()~\cite{dettmers2022gpt3}, NF4 from QLORA~\cite{dettmers2024qlora}, and SVDQuant~\cite{li2024svdquant} methods.
All similarities were computed using 10,000 DiffusionDB prompts, similar to Figure~\ref{fig:semantic_drift_general}.

\begin{figure}[ht]
    \centering
    \begin{subfigure}{0.32\textwidth}
        \centering
        \includegraphics[width=\linewidth]{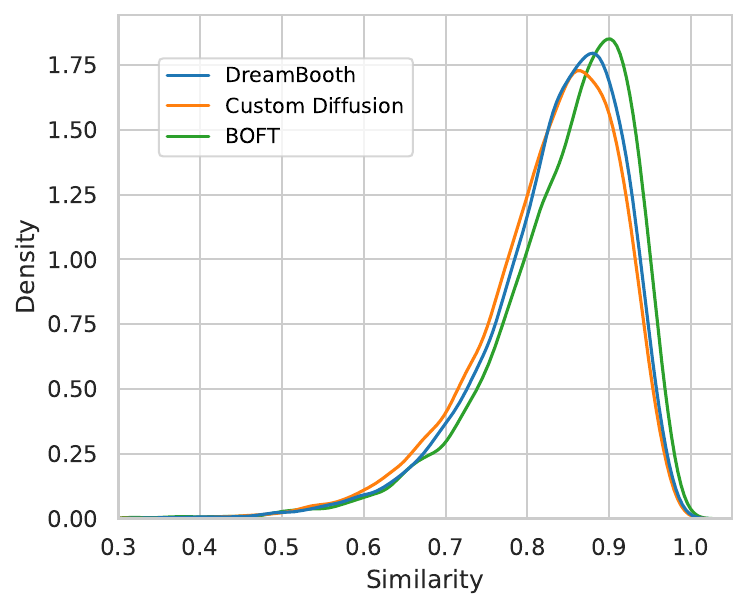}
        \caption{Customization}
        \label{fig:areas_forgetting:customization}
    \end{subfigure}
    \begin{subfigure}{0.32\textwidth}
        \centering
        \includegraphics[width=\linewidth]{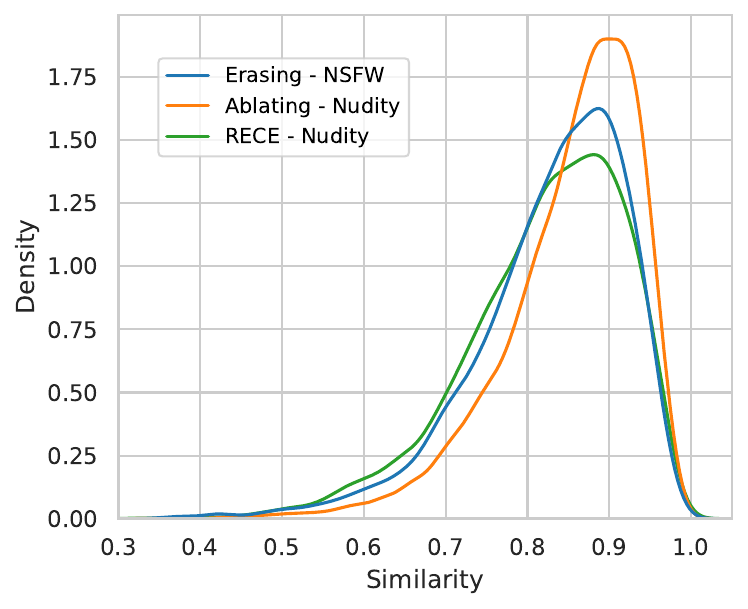}
        \caption{Unlearning}
        \label{fig:areas_forgetting:unlearning}
    \end{subfigure}
    \begin{subfigure}{0.32\textwidth}
        \centering
        \includegraphics[width=\linewidth]{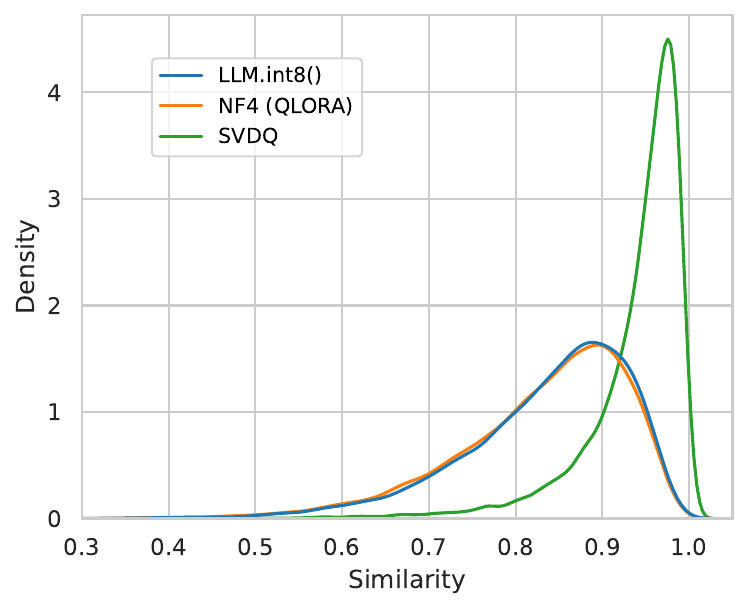}
        \caption{Quantization}
        \label{fig:areas_forgetting:quantization}
    \end{subfigure}
    \caption{Distribution of CLIP-I cosine similarities between base model and modified model outputs across different modification approaches. All approaches exhibit significant and similar shifting behavior.}
    \label{fig:areas_forgetting}
\end{figure}

As shown in Figure~\ref{fig:areas_forgetting}, all modification approaches exhibit significant distributional shifts from the base model's behavior, showing similar distributions. This suggests comparable levels of forgetting despite their different adaptation approaches, with SVDQuant showing a particularly sharp peak near similarity 0.95, indicating more consistent but still present forgetting effects.
These results reinforce our hypothesis that open-world forgetting is inherent to weight modification, regardless of the intended purpose or method of modification.

\subsection{Large-scale Concept Evaluation} \label{sec:exp_scaling}

To assess the generalizability of our findings, we conducted an experiment using all concepts from CustomConcept101. We evaluated zero-shot classification performance through multiple random sampling runs for each concept count N (1 to 100), computing confidence intervals for statistical reliability. For each dataset and N value, we performed 10 independent sampling runs.

\begin{figure}[ht]
    \centering
    \includegraphics[width=0.6\linewidth]{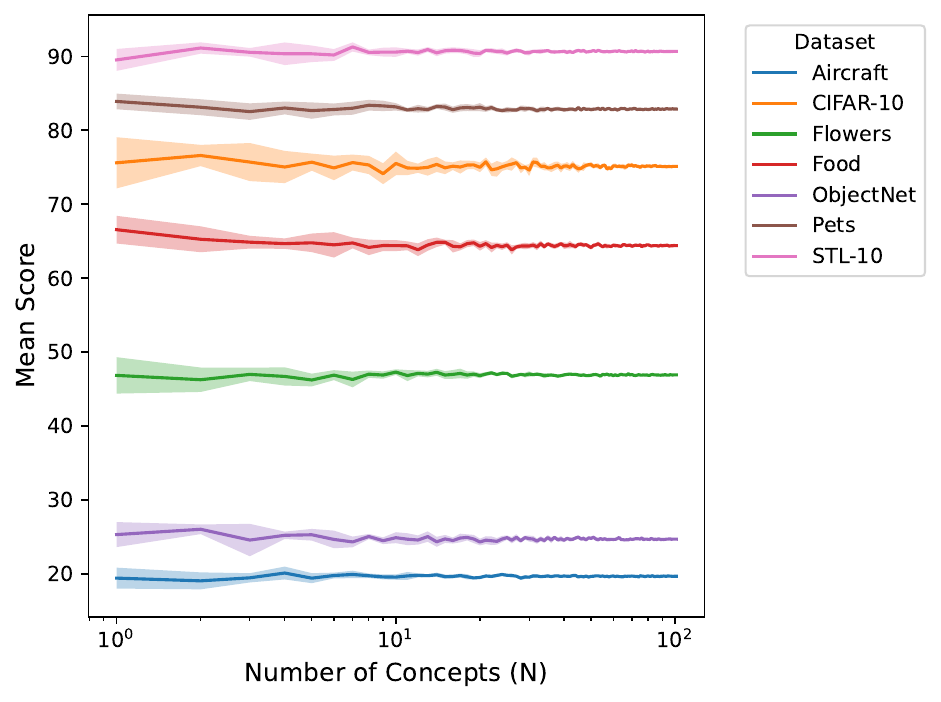}
    \caption{Zero-shot classification performance as a function of the number of concepts used in evaluation. The stability across all datasets suggests that semantic drift findings generalize beyond the small-scale evaluations.}
    \label{fig:exp_scaling}
\end{figure}

Figure~\ref{fig:exp_scaling} shows the mean classification score in seven datasets (Aircraft, CIFAR-10, Flowers, Food, ObjectNet, Pets, and STL-10), plotting the mean classification score against the number of concepts on a logarithmic scale. Demonstrating consistent classification performance across different scales, persisting across datasets with varying baseline performance levels, from STL-10 ($\sim 90\%$) and Pets ($\sim 80\%$) to Aircraft ($\sim 20\%$).

The nearly horizontal trend lines and narrow confidence intervals suggest our findings about representation drifting are scale-invariant. This implies that insights from smaller-scale evaluations can reliably generalize to larger concept sets.

\subsection{Cross-Architecture Analysis} \label{sec:model_families}

\begin{figure}[ht]
    \centering
    \includegraphics[width=0.5\linewidth]{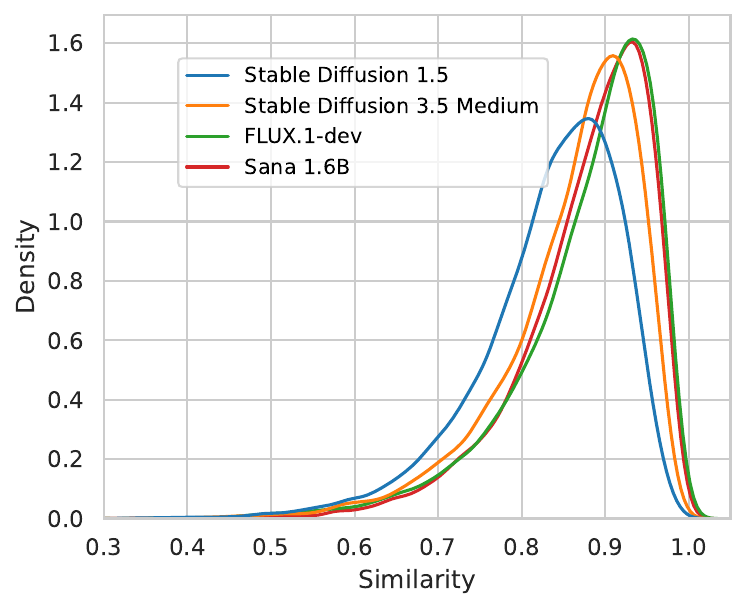}
    \caption{CLIP-I similarity distributions between base and DreamBooth-modified versions across different model architectures. All show characteristic shifts in their output distribution.}
    \label{fig:model_families}
\end{figure}

To investigate whether open-world forgetting is specific to particular architectures, we applied DreamBooth to different families of generative models: Stable Diffusion 1.5~\cite{Rombach_2022_CVPR_stablediffusion}, Stable Diffusion 3.5 Medium~\cite{esser2024scaling}, FLUX.1 [dev]~\cite{flux2023}, and Sana 1.6B~\cite{xie2024sana}. Despite their architectural differences, all models exhibit similar distributional shifts when modified through DreamBooth (Figure~\ref{fig:model_families}). The rightward skew in the similarity distributions suggests that while the degree of forgetting may vary slightly between architectures, the fundamental phenomenon persists across model families. This provides evidence that open-world forgetting emerges naturally from weight adjustments in large-scale generative models, rather than being a byproduct of particular architectural decisions.

\section{Additional Results} \label{sec:additional_results}

\subsection{Semantic Drift} \label{sec:additional_semantic}

In this section we show additional methods for Table~\ref{tab:diffusion_classifier_and_concept_evaluation} as well as standard deviations for statistical significance. 

\begin{table}[ht]
    \caption{Zero-shot classification using the T2I model. Personalized models suffer from degraded representations. Worst class drop between parenthesis. Scores with standard deviation across models.}
    \label{tab:appx:diffusion_classifier_stds}
    \begin{center}
    \begin{tabular}{@{}lcccccccc@{}}
    \toprule
            & Food                           & CIFAR10 & Aircraft & Pets & Flowers & STL10 & ObjectNet \\ \midrule
    Base Model & 71.09 & 81.60 & 23.40 & 86.87 & 50.00 & 93.00 & 28.50 \\
    DB FT & 61.50{\scriptsize $\pm$5.95} & 69.86{\scriptsize $\pm$6.79} & 16.04{\scriptsize $\pm$3.81} & 79.11{\scriptsize $\pm$5.29} & 43.18{\scriptsize $\pm$6.58} & 87.04{\scriptsize $\pm$4.63} & 21.52{\scriptsize $\pm$3.41} \\
    DB & 65.48{\scriptsize $\pm$2.59} & 75.92{\scriptsize $\pm$5.81} & 19.36{\scriptsize $\pm$2.69} & 82.61{\scriptsize $\pm$3.33} & 46.61{\scriptsize $\pm$2.49} & 91.30{\scriptsize $\pm$2.05} & 25.26{\scriptsize $\pm$1.85} \\
    DB-DC & 69.07{\scriptsize $\pm$1.58} & 80.98{\scriptsize $\pm$2.57} & 21.42{\scriptsize $\pm$0.45} & 86.64{\scriptsize $\pm$0.92} & 49.29{\scriptsize $\pm$1.33} & 93.36{\scriptsize $\pm$0.70} & 27.72{\scriptsize $\pm$1.56} \\
    DB-DC$\backslash$pr & 68.79{\scriptsize $\pm$1.78} & 80.60{\scriptsize $\pm$2.46} & 21.54{\scriptsize $\pm$0.77} & 86.37{\scriptsize $\pm$0.78} & 49.06{\scriptsize $\pm$1.12} & 92.94{\scriptsize $\pm$0.95} & 27.45{\scriptsize $\pm$1.15} \\
    CD & 65.25{\scriptsize $\pm$2.76} & 79.98{\scriptsize $\pm$4.21} & 19.44{\scriptsize $\pm$1.97} & 83.46{\scriptsize $\pm$2.97} & 47.65{\scriptsize $\pm$2.34} & 91.40{\scriptsize $\pm$1.60} & 25.75{\scriptsize $\pm$2.16} \\
    CD-DC & 69.19{\scriptsize $\pm$1.73} & 82.36{\scriptsize $\pm$1.91} & 21.94{\scriptsize $\pm$1.51} & 86.37{\scriptsize $\pm$1.28} & 49.33{\scriptsize $\pm$1.16} & 93.02{\scriptsize $\pm$0.97} & 27.91{\scriptsize $\pm$1.30} \\
    CD-DC$\backslash$pr & 68.99{\scriptsize $\pm$1.73} & 82.04{\scriptsize $\pm$2.23} & 21.56{\scriptsize $\pm$1.94} & 86.70{\scriptsize $\pm$1.22} & 49.16{\scriptsize $\pm$1.31} & 92.76{\scriptsize $\pm$0.85} & 27.77{\scriptsize $\pm$1.58} \\
    SVDiff & 77.40{\scriptsize $\pm$2.32} & 91.45{\scriptsize $\pm$1.72)} & 46.94{\scriptsize $\pm$1.29} & 84.66{\scriptsize $\pm$1.01} & 25.88{\scriptsize $\pm$1.95} & 65.82{\scriptsize $\pm$2.78} & 19.32{\scriptsize $\pm$0.76} \\
    SVDiff-DC & 80.60{\scriptsize $\pm$1.40} & 92.95{\scriptsize $\pm$0.50} & 48.35{\scriptsize $\pm$1.10} & 85.69{\scriptsize $\pm$0.71} & 26.77{\scriptsize $\pm$1.39} & 69.27{\scriptsize $\pm$1.62} & 20.04{\scriptsize $\pm$0.68} \\
    BOFT & 76.60{\scriptsize $\pm$4.76} & 89.75{\scriptsize $\pm$5.20} & 46.04{\scriptsize $\pm$2.39} & 82.86{\scriptsize $\pm$1.93} & 25.26{\scriptsize $\pm$2.10} & 64.79{\scriptsize $\pm$1.90} & 20.05{\scriptsize $\pm$1.98} \\
    BOFT-DC & 78.72{\scriptsize $\pm$1.29} & 91.20{\scriptsize $\pm$1.75} & 48.43{\scriptsize $\pm$1.90} & 85.44{\scriptsize $\pm$1.17} & 26.20{\scriptsize $\pm$1.54} & 67.33{\scriptsize $\pm$0.58} & 19.35{\scriptsize $\pm$1.84} \\
    \bottomrule
    \end{tabular}
    \end{center}
\end{table}

\begin{figure}[ht]
    \centering
    \includegraphics[height=0.95\textheight]{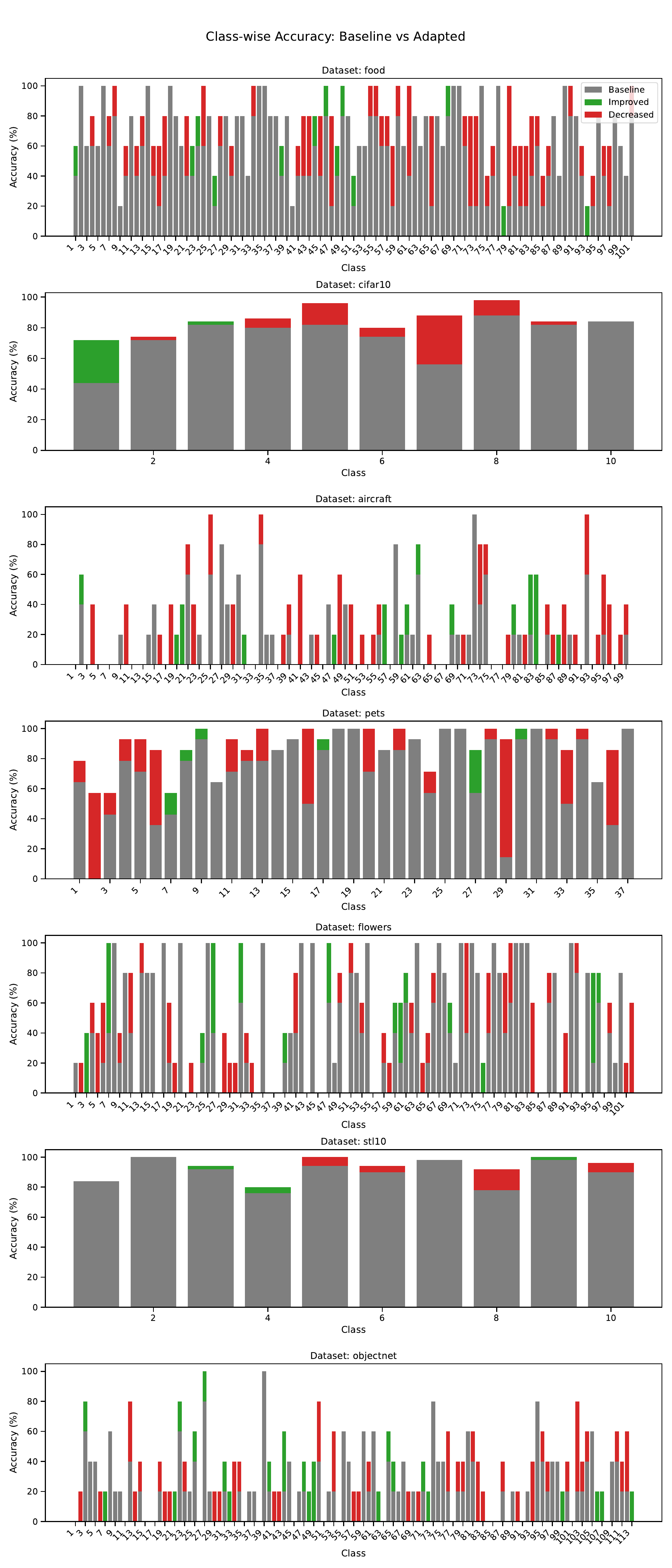}
    \caption{Class-wise accuracy of base model and DreamBooth adaptation to the ``decoritems\_lamp1'' concept for several data sets.}
    \label{fig:appx:classwise_accuracy}
\end{figure}

As further illustration in Figure~\ref{fig:appx:classwise_accuracy} of the zero-shot classification accuracy, we provide the results for one of the models, namely ``decoritems\_lamp1'' for all the datasets for DreamBooth. We can observe that the adaptation leads to drops on most classes (identified in red) but can also occasionally result in a performance increase (in green). 

\begin{table}[ht]
    \caption{Concept fidelity (DINO, CLIP-I) and prompt fidelity (CLIP-T). Drift Correction maintains fidelity across metrics.}
    \label{tab:appx:concept_evaluation_stds}
    \centering    
    \begin{tabular}{@{}lccc@{}}
    \toprule
            & DINO & CLIP-I & CLIP-T \\ \midrule
    DB & 0.4241{\scriptsize $\pm$0.1503} & 0.6764{\scriptsize $\pm$0.1046} & 0.7896{\scriptsize $\pm$0.0296} \\
    DB-DC & 0.4283{\scriptsize $\pm$0.1584} & 0.6817{\scriptsize $\pm$0.1097} & 0.7799{\scriptsize $\pm$0.0324} \\
    DB-DC$\backslash$pr & 0.4315{\scriptsize $\pm$0.1585} & 0.6841{\scriptsize $\pm$0.1086} & 0.7776{\scriptsize $\pm$0.0322} \\
    CD & 0.4422{\scriptsize $\pm$0.1378} & 0.6934{\scriptsize $\pm$0.0902} & 0.7916{\scriptsize $\pm$0.0266} \\
    CD-DC & 0.4381{\scriptsize $\pm$0.4381} & 0.6925{\scriptsize $\pm$0.0888} & 0.7899{\scriptsize $\pm$0.0280} \\
    CD-DC$\backslash$pr & 0.4382{\scriptsize $\pm$0.1351} & 0.6935{\scriptsize $\pm$0.0872} & 0.7872{\scriptsize $\pm$0.0264} \\
    SVDiff & 0.5017{\scriptsize $\pm$0.1404} & 0.7493{\scriptsize $\pm$0.0578} & 0.7442{\scriptsize $\pm$0.0497} \\
    SVDiff-DC & 0.4288{\scriptsize $\pm$0.1804} & 0.6960{\scriptsize $\pm$0.1144} & 0.7287{\scriptsize $\pm$0.0470} \\
    BOFT & 0.3765{\scriptsize $\pm$0.1797} & 0.6674{\scriptsize $\pm$0.0983} & 0.7547{\scriptsize $\pm$0.0291} \\
    BOFT-DC & 0.4043{\scriptsize $\pm$0.1789} & 0.6705{\scriptsize $\pm$0.0991} & 0.7618{\scriptsize $\pm$0.0341} \\
    \bottomrule
    \end{tabular}
\end{table}

\subsection{Appearance Drift} \label{appx:appearance_drift_full}
See Tables \ref{tab:appereance_metric_comparison_simple} and \ref{tab:appereance_metric_comparison} for the full results which have been used for the generation of Figure~\ref{fig:cdi_radar}. The prompt are `0:Face',`1:Pedestrian',`2:Car',`3:Cow',`4:Hand',`5:Chair',`6:Mountain', `7:Beach',`8:Forest',`9:Highway',`10:Street',`11:Indoor', 
`12:Animal in natural scene', `13:Tree in urban scene', `14:Close-up person in urban scene',`15:Far pedestrian in urban scene',`16:Car in urban scene',`17:Lamp in indoor scene', `18:empty prompt'.

\begin{table}[ht]
\caption{Comparison of CDI, FID, and KID Values for custom vs custom-regularized methods (similar to those in Figure~\ref{fig:cdi_radar}). }
\label{tab:appereance_metric_comparison_simple}
\centering
\begin{tabular}{lccc}
\toprule
Method & CDI $\downarrow$ & KID $\downarrow$ & FID $\downarrow$ \\
\midrule
Base Model & 0.11{\scriptsize $\pm$0.03} & 0.01{\scriptsize $\pm$0.01} & 31.32{\scriptsize $\pm$10.15} \\
DB & 0.56{\scriptsize $\pm$0.19} & 1.41{\scriptsize $\pm$0.82} & 45.97{\scriptsize $\pm$12.39} \\
DB-DC & 0.30{\scriptsize $\pm$0.11} & 0.27{\scriptsize $\pm$0.19} & 29.07{\scriptsize $\pm$7.59} \\
CD & 0.52{\scriptsize $\pm$0.13} & 1.40{\scriptsize $\pm$0.70} & 46.91{\scriptsize $\pm$12.89} \\
CD-DC & 0.30{\scriptsize $\pm$0.07} & 0.42{\scriptsize $\pm$0.35} & 33.06{\scriptsize $\pm$12.26} \\
SVDiff & 0.87{\scriptsize $\pm$0.14} & 1.15{\scriptsize $\pm$0.67} & 52.55{\scriptsize $\pm$12.95} \\
SVDiff-DC & 0.30{\scriptsize $\pm$0.03} & 0.39{\scriptsize $\pm$0.22} & 45.64{\scriptsize $\pm$12.74} \\
BOFT & 0.57{\scriptsize $\pm$0.11} & 1.12{\scriptsize $\pm$0.72} & 52.66{\scriptsize $\pm$18.94} \\
BOFT-DC & 0.34{\scriptsize $\pm$0.05} & 0.36{\scriptsize $\pm$0.20} & 40.82{\scriptsize $\pm$10.04} \\
\bottomrule
\end{tabular}
\end{table}

\begin{table}[ht]
\caption{Comparison of CDI, FID, and KID Values for custom vs custom-regularized methods for the prompts (similar to those  in Figure~\ref{fig:cdi_radar}). }
\label{tab:appereance_metric_comparison}
\centering
\tiny
\begin{tabular}{cccccccccccccccc}
\toprule
{Prompt} & \multicolumn{3}{c}{Vanilla} & \multicolumn{3}{c}{DB} & \multicolumn{3}{c}{DB-DC} & \multicolumn{3}{c}{CD} & \multicolumn{3}{c}{CD-DC} \\
\cmidrule(lr){2-4} \cmidrule(lr){5-7} \cmidrule(lr){8-10} \cmidrule(lr){11-13} \cmidrule(lr){14-16}
 & CDI & FID & KID & CDI & FID & KID & CDI & FID & KID & CDI & FID & KID & CDI & FID & KID \\
\midrule
00 & 0.10 & 24.75 & 0.00 & 0.65 & 43.74 & 2.00 & 0.23 & 21.70 & 0.16 &  0.48 & 37.01 & 1.22 & 0.24 & 24.90 & 0.29 \\
01 & 0.14 & 44.55 & 0.00 & 1.01 & 79.78 & 3.22 & 0.43 & 46.16 & 0.50 & 0.82 & 72.44 & 2.63 & 0.33 & 51.68 & 0.79 \\
02 & 0.12 & 19.70 & 0.01 & 0.35 & 28.31 & 0.90 & 0.18 & 18.37 & 0.12 & 0.31 & 27.14 & 0.58 & 0.19 & 20.11 & 0.15 \\
03 & 0.09 & 29.17 & 0.01 & 0.36 & 41.81 & 1.22 & 0.14 & 27.40 & 0.13 & 0.40 & 42.25 & 1.13 & 0.23 & 31.62 & 0.37 \\
04 & 0.17 & 35.93 & 0.01 & 0.59 & 40.77 & 0.59 & 0.23 & 30.53 & 0.07 & 0.57 & 42.65 & 0.70 & 0.32 & 33.76 & 0.22 \\
05 & 0.14 & 19.65 & 0.01 & 0.76 & 28.15 & 0.61 & 0.27 & 17.93 & 0.08 &  0.58 & 26.45 & 0.57 & 0.34 & 20.09 & 0.17 \\
06 & 0.09 & 34.22 & 0.00 & 0.48 & 52.80 & 1.46 & 0.26 & 34.55 & 0.32 & 0.57 & 60.90 & 2.16 & 0.34 & 41.52 & 0.72 \\
07 & 0.14 & 34.89 & 0.05 & 0.36 & 42.58 & 0.64 & 0.25 & 29.45 & 0.14 & 0.41 & 48.98 & 1.19 & 0.22 & 33.66 & 0.26 \\
08 & 0.10 & 25.50 & 0.00 & 0.48 & 52.42 & 2.79 & 0.25 & 26.78 & 0.57 & 0.42 & 55.33 & 2.95 & 0.29 & 32.76 & 1.01 \\
09 & 0.12 & 27.51 & 0.01 & 0.51 & 41.82 & 1.73 & 0.30 & 28.22 & 0.30 & 0.43 & 41.37 & 1.40 & 0.33 & 27.80 & 0.52 \\
10 & 0.10 & 24.34 & 0.00 & 0.62 & 47.28 & 2.78 & 0.34 & 26.27 & 0.68 & 0.69 & 44.67 & 2.09 & 0.42 & 28.39 & 0.72 \\
11 & 0.06 & 33.56 & 0.02 & 0.38 & 46.44 & 1.39 & 0.19 & 30.86 & 0.27 & 0.53 & 52.92 & 1.86 & 0.26 & 35.79 & 0.53 \\
12 & 0.11 & 29.75 & 0.01 & 0.43 & 43.92 & 1.20 & 0.24 & 28.67 & 0.21 & 0.41 & 49.51 & 1.69 & 0.27 & 36.02 & 0.68 \\
13 & 0.13 & 20.26 & 0.00 & 0.66 & 44.59 & 1.50 & 0.64 & 20.36 & 0.47 & 0.62 & 43.14 & 1.40 & 0.42 & 30.33 & 0.37 \\
14 & 0.12 & 40.60 & 0.01 & 0.53 & 47.32 & 0.58 & 0.33 & 36.76 & 0.16 & 0.60 & 48.05 & 0.66 & 0.33 & 38.68 & 0.18 \\
15 & 0.07 & 36.90 & 0.01 & 0.96 & 60.22 & 1.96 & 0.48 & 38.81 & 0.51 & 0.75 & 59.16 & 1.92 & 0.46 & 40.97 & 0.51 \\
16 & 0.09 & 26.65 & 0.00 & 0.68 & 37.07 & 1.02 & 0.41 & 26.12 & 0.26 & 0.66 & 36.38 & 0.90 & 0.41 & 28.21 & 0.29 \\
17 & 0.13 & 27.73 & 0.01 & 0.59 & 32.44 & 0.67 & 0.30 & 25.44 & 0.23 & 0.52 & 33.93 & 0.74 & 0.27 & 25.77 & 0.14 \\
18 & 0.18 & 59.52 & 0.02 & 0.29 & 62.02 & 0.59 & 0.23 & 37.99 & 0.00 & 0.31 & 68.52 & 0.93 & 0.21 & 48.00 & 0.13 \\
\bottomrule
\end{tabular}
\end{table}

\section{Ablations} \label{sec:ablations}

\subsection{Does finetuning lead to loss of diversity?}

Finetuning large foundational models on a limited set of images (typically around 5) of a specific subject can lead to overfitting, a phenomenon observed in previous studies such as DreamBooth. This overfitting often results in a loss of diversity in generated images and a noticeable shift towards the characteristics of the training subject. While prior regularization techniques have been employed to mitigate this shift, they have not fully resolved the issue, as our analysis demonstrates.

To assess the impact of finetuning on diversity, we adapt the metric introduced in the DreamBooth study. This metric quantifies diversity by calculating the average Learned Perceptual Image Patch Similarity (LPIPS) between generated images of the same subject using identical prompts. A higher LPIPS score indicates greater diversity among the generated images.

Our proposed method not only improves the mitigation of subject shifting, as evidenced in Figures~\ref{fig:cdi_radar} and \ref{fig:semantic_drift_superclass_general}, but also maintains the diversity of the original model. To validate this, we conducted an extensive evaluation using 100 different prompts, each generating 100 images. These prompts were sourced from the DiffusionDB subset, as detailed in Appendix~\ref{appx:drift_exp_setup}.

Figure~\ref{fig:appx:diversity} presents the results of our diversity analysis. The data demonstrates that our method preserves diversity at a level comparable to, or even exceeding, previous approaches. This finding is particularly significant as it indicates that our technique not only addresses the shifting problem more effectively but does so without compromising the model's ability to generate diverse outputs.

The preservation of diversity while improving subject fidelity represents a crucial advancement in finetuning methodologies for large generative models. It ensures that the fine-tuned model retains its creative capacity and versatility across a wide range of prompts and subjects, even as it gains enhanced capabilities in representing specific training subjects. This balance between specificity and diversity is essential for the practical application of fine-tuned models in various creative and technical domains.

\begin{figure}[ht]
    \centering
    \includegraphics[width=0.8\linewidth]{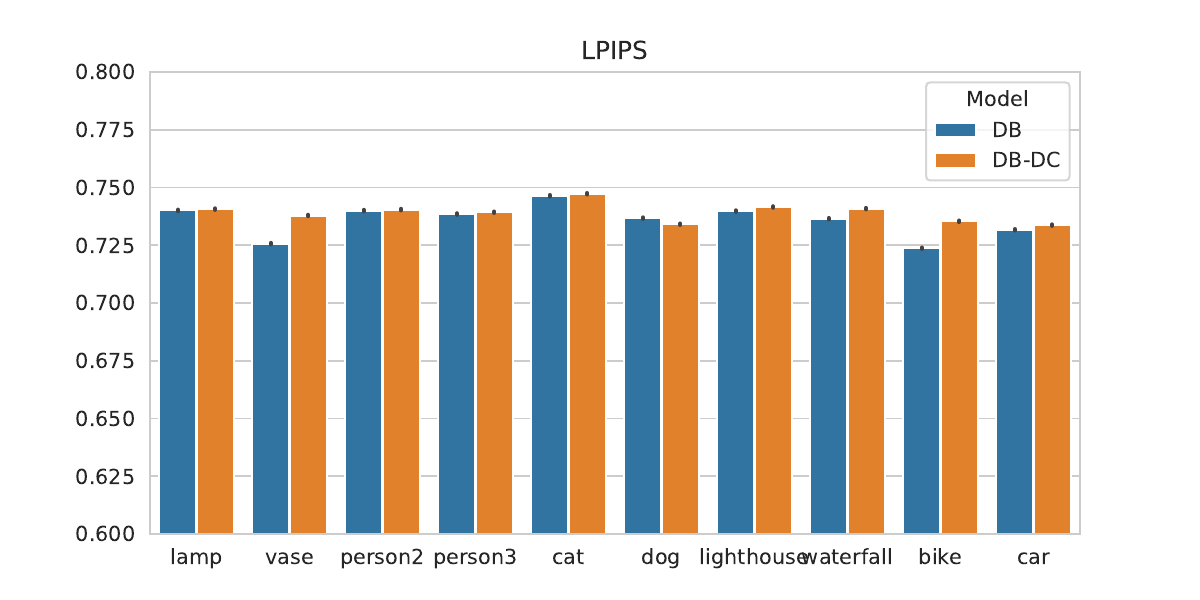}
    \caption{Diversity evaluation using LPIPS. Drift Correction maintains the original diversity after adaptation.}
    \label{fig:appx:diversity}
\end{figure}

\subsection{Increasing the buffer size reduces drift}

To investigate the impact of buffer size on mitigating open-world forgetting, we conducted experiments varying the number of images in the replay buffer during model adaptation. Table~\ref{tab:ablation:buffer_size} presents the results of this analysis, showing the effect of buffer size on both semantic drift (measured by zero-shot CIFAR10 classification accuracy) and appearance drift (measured by Color Drift Index, CDI).

\begin{table}[ht]
    \centering
    \caption{Effect of the number of images in the buffer.}
    \label{tab:ablation:buffer_size}
    \begin{tabular}{@{}lccccccc@{}}
        \toprule
        Metric               & 0 & 50 & 100 & 200 & 500 & 1000 & 2000 \\ \midrule
        Acc {\scriptsize CIFAR10} & 63.24{\scriptsize $\pm$14.61} & 76.10{\scriptsize $\pm$6.66} & 75.62{\scriptsize $\pm$6.48} & 75.92{\scriptsize $\pm$5.81} & 76.36{\scriptsize $\pm$5.28} & 76.92{\scriptsize $\pm$5.87} & 77.12{\scriptsize $\pm$6.20} \\
        CDI                 & 0.87  &  0.64  &  0.58   &  0.56   &  0.70   &   0.60   &  0.51    \\ \bottomrule
    \end{tabular}
\end{table}

The experiment suggests that incorporating a replay buffer, even of modest size, is beneficial for mitigating open-world forgetting, particularly in terms of semantic drift. However, the benefits of increasing buffer size show diminishing returns, especially for semantic preservation. For appearance drift, while there is a general trend towards improvement with larger buffers, significant drift persists regardless of buffer size.

\subsection{Influence of the training images}

\begin{figure}[ht]
    \centering
    \includegraphics[width=1\linewidth]{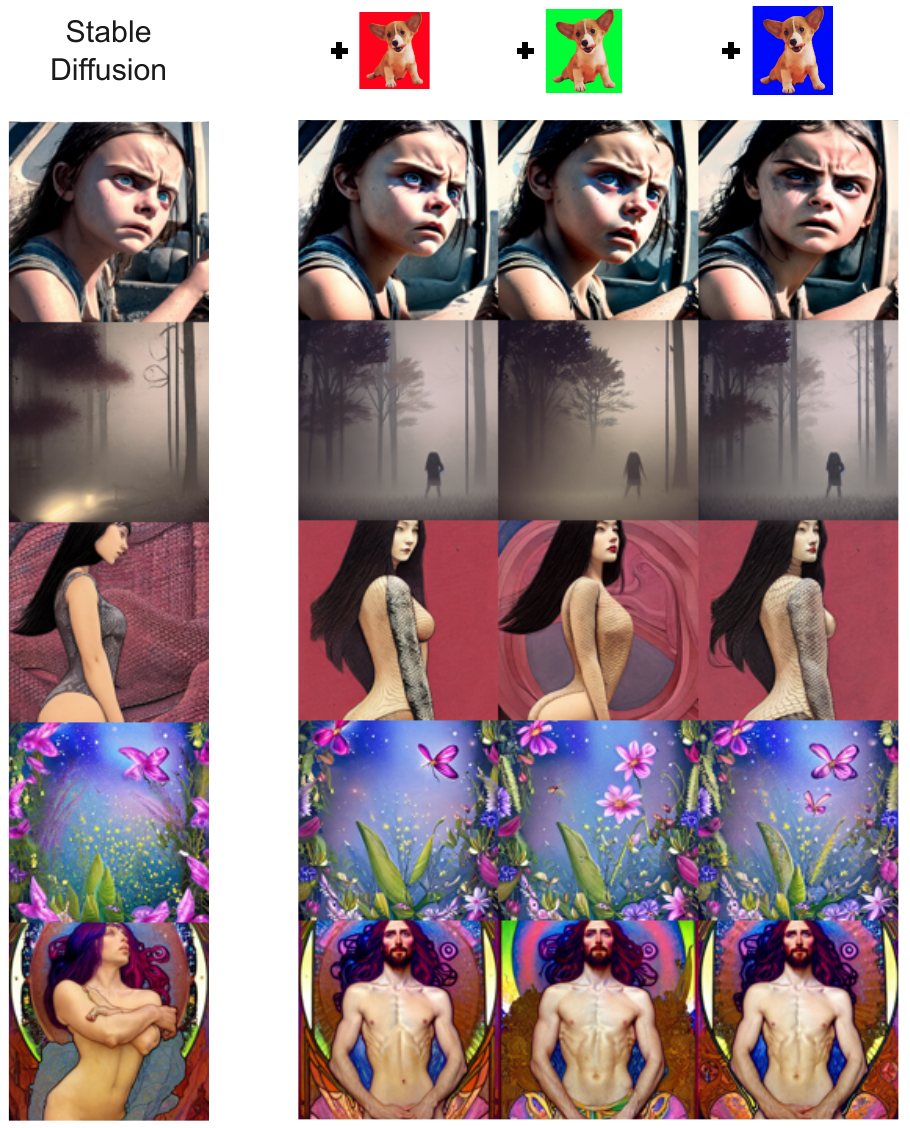}
    \caption{Appearance drift variation as a function of background color in training samples}
    \label{fig:appx:color_buffer}
\end{figure}

Our experiments reveal that the characteristics of the training images used during model adaptation can significantly impact the nature and extent of appearance drift. To illustrate this effect, we conducted an experiment in Figure~\ref{fig:appx:color_buffer} focusing on how the background color in training samples influences the color distribution of generated images. The results show that the background color of the training images has a noticeable impact on the color distribution of the generated images, even when generating images of unrelated concepts.

These findings highlight the importance of carefully considering the visual characteristics of training images when adapting generative models. The background, lighting, and overall composition of training samples can have far-reaching effects on the model's output distribution, extending beyond the specific concept being learned. 

\section{User Study} \label{sec:user_study}
We conducted a user study with 24 participants, evaluating 100 image pairs (before/after customization) across 20 concepts for both DreamBooth and Custom Diffusion. In each trial, participants were asked ``Which image is the most similar to the center image?'' where the center image was generated by the base model, and the left and right images (randomly assigned) were from the customized model and the customized model with our drift correction method. 

\begin{figure}[ht]
    \centering
    \includegraphics[width=0.5\linewidth]{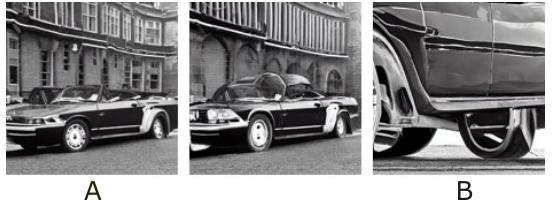}
    \caption{Example triad presented in the user study. }
    \label{fig:appx:user_study}
\end{figure}

In the user study, participants were presented with image triads, as shown in Figure~\ref{fig:appx:user_study}. Each triad consisted of a reference image in the center and two comparison images (labeled A and B) on either side. The methods evaluated were DreamBooth and Custom Diffusion and its corresponding Drift Corrected versions. Participants were given the following instructions:

\textit{"Look at the three images shown: one in the center, and two options (A and B) on the sides. 
Your task is to determine which side image (A or B) is more visually similar to the center image."}

The results strongly favor our approach, with participants preferring our drift-corrected outputs 66\% of the time compared to baseline customization. The standard deviation across users was just 3.95\%, and statistical analysis confirms the significance of these results (paired t-test: $t(23)=19.624$, $p < 0.001$; Wilcoxon signed-rank test: $p < 0.001$). These results demonstrate consistent preference for our method across all participants and validate our quantitative metrics with human perception.

\section{Experiment Details} \label{appx:exp_details}

This section outlines our experimental setup, including datasets, metrics, and training configurations.

\subsection{Semantic Drift Evaluation} \label{appx:drift_exp_setup}

\paragraph{Datasets.} To evaluate open-world forgetting, we select a random subset of 10,000 user prompts from DiffusionDB 2M~\citep{wangDiffusionDBLargescalePrompt2023} (Table~\ref{tab:appx:diffusiondb_samples}).
For adaptation training and evaluation, we choose a subset of 10 concepts from CustomConcept101~\citep{kumari2022customdiffusion}, namely \textit{decoritems\_lamp1}, \textit{decoritems\_vase2}, \textit{person\_2}, \textit{person\_3}, \textit{pet\_cat5}, \textit{pet\_dog4}, \textit{transport\_bike}, \textit{transport\_car2}, \textit{scene\_lighthouse}, \textit{scene\_waterfall}. Each concept contains approximately 3-5 images.
For superclass evaluation, we create a dataset of 10 synonyms with respect to each concept, which can be found in Table~\ref{appx:tab:concept_synonyms}.

\begin{table}[ht]
\caption{DiffusionDB subset sample prompts. Shorter prompts selected for visualization purposes.}
\label{tab:appx:diffusiondb_samples}
\resizebox{\textwidth}{!}{%
\begin{tabular}{@{}c@{}}
\toprule
DiffusionDB prompts                                                                                                                                               \\ \midrule
``\textit{dafne keen, mad max, cinematic shot, 8k resolution}''                                                                                                   \\
``\textit{creepy horror movie characters, fog, rain, volumetric lighting, beautiful, golden hour, sharp focus, highly detailed, cgsociety}''                      \\
``\textit{the railroad is a place of death. it's where the forgotten and the damned go to die. it's a place of dark secrets and hidden terror. photorealistic }'' \\
``\textit{samurai jack johnny bravo by salvador dali }''                                                                                                          \\
``\textit{Film still of Emma Watson as Princess Leia in Star Wars (1977)}''                                                                                       \\
``\textit{a detailed figure of indigo montoya, first 4 figures, detailed product photo }''                                                                        \\
``\textit{a hyper scary pokemon, horror, creepy, big budget horror movie, by zdzisław beksinski, by dorian cleavenger }''                                         \\
``\textit{the war between worlds extremely detailed claymation art, dark, moody, foggy }''                                                                        \\
``\textit{a painting of Hatsune Miku by H. R. Giger, highly detailed, 4k digital art}''                                                                           \\
``\textit{a redneck with wings and horns wearing sunglasses and snake skin smoking a blunt, detailed, 4 k, realistic, picture }''                                 \\
``\textit{fantasy art 4 k ultra detailed photo caricature walter matthau as an fighter pilot }''                                                                  \\
``\textit{ CG Homer Simpson as Thanos, cinematic, 4K}''                                                                                                           \\
``\textit{Full body portrait of Raven from Teen Titans (2003), digital art by Sakimichan, trending on ArtStation}''                                               \\
``\textit{bigfoot walking down the street in downtown Bremerton Washington}''                                                                                     \\
``\textit{garden layout rendering with flowers and plants native to ottawa canada }''                                                                             \\
``\textit{a beautiful planet of guangzhou travel place of interest, chill time. good view, exciting honor. by david inshaw }''                                    \\
``\textit{an oil painting of Dwayne Johnson instead of Mona Lisa in the famous painting The Joconde painted by Leonardo Da Vinci}''                               \\
``\textit{film still of danny devito as mario in live action super mario bros movie, 4 k }''                                                                      \\
``\textit{a beautiful artist's rendition of what the stable diffusion algorithm dreams about }''                                                                  \\ \bottomrule
\end{tabular}
}
\end{table}

\begin{table}[ht]
\caption{Concept synonyms.}
\label{appx:tab:concept_synonyms}
\resizebox{\textwidth}{!}{
\begin{tabular}{@{}cc@{}}
\toprule
Concept    & Synonyms                                                                                                              \\ \midrule
bike       & pedal cycle, velociped, roadster, bicycle, push bike, pushbike, cycle,wheels, two-wheeler, pedal bike                 \\
car        & jalopy, ride, auto, vehicle, coupe, wheels, automobile, sedan, hatchback, motocar                                     \\
cat        & feline, grimalkin, mouser, moggy, tabby, puss, kitty, kitten, pussycat, tomcat                                        \\
dog        & canine, hound, pup, pooch, fido, puppy, mutt, man's best friend, doggy, cur                                           \\
lamp       & fixture, chandelier, light, illuminator, lantern, luminaire, glow, torch, sconce, beacon                              \\
lighthouse & light, coastal beacon, navigation light, pharos, seamark, watchover, beacon, guide light, light station, signal tower \\
person     & gent, bloke, chap, gentleman, lad, guy, male, bro, fellow, dude                                                       \\
vase       & urn, amphora, container, pitcher, carafe, receptacle, jar, vessel, pot, jug                                           \\
waterfall  & rapids, torrent, flume, cascade, spillway, cataract, plunge, chute, falls, deluge                                     \\ \bottomrule
\end{tabular}
}
\end{table}

\paragraph{Metrics.} We employ three primary metrics to assess image generation quality. CLIP-I is calculated as the average pairwise cosine similarity between CLIP \citep{radford2021clip} embeddings of real and generated images. DINO uses the same pairwise cosine similarity method but with DINO \citep{caron2021emerging} ViT-S16 embeddings. This metric is preferred over CLIP-I as it does not ignore differences between subjects of the same class. CLIP-T measures the CLIP embedding cosine similarity between the prompt and the generated image, and is used to evaluate prompt fidelity.

\paragraph{Training configuration.} We adapt models using publicly available scripts from Diffusers~\citep{von-platen-etal-2022-diffusers} for Dreambooth\footnote{\url{https://github.com/huggingface/diffusers/blob/main/examples/dreambooth/train_dreambooth_lora.py}} and Custom Diffusion\footnote{\url{https://github.com/huggingface/diffusers/blob/main/examples/custom_diffusion/train_custom_diffusion.py}} applied to Stable Diffusion v1.5 \citep{Rombach_2022_CVPR_stablediffusion}. Both methods use prior regularization unless otherwise stated, which is designed to prevent drifting towards the training concept. The set of images for prior regularization is generated from the base model before training starts. 
We use the LoRA script versions and refer to the resulting models as DB for DreamBooth and CD for Custom Diffusion. Full finetuning models exhibit the same or worse shortcomings as the LoRA models analyzed and are termed FT for finetuning, such as DB FT in Table~\ref{tab:appx:diffusion_classifier_stds}.
Since DB and CD are similar and to ensure a fair comparison, both methods use similar training settings: a learning rate of 1e-4, batch size of 1, 500 training steps, and no augmentations. The prior regularization uses a weighting of 1 and comprises 200 samples of generated images with the prompt ``\{concept\}'' for each concept, using default generation settings. For the drift correction method described in Section~\ref{sec:drift_correction}, all settings remain the same, and the weighting parameter is set to $\lambda = 10$.

\subsection{Diffusion Classifier} \label{appx:diffusion_classifier}

We employ the official released code of the Diffusion Classifier method\footnote{\url{https://github.com/diffusion-classifier/diffusion-classifier}}. However, due to computational constraints, we modify some parameters for our explorations. We reduce the keep list to $(10, 100)$ across all datasets while maintaining the trial list at $(5, 1)$. This significantly reduces computational time while resulting in minimal percentual score uncertainty. 
Additionally, datasets with many classes or samples were reduced to have a total number of samples of roughly 500 by random selection of the samples of each class. The datasets configuration can be seen in Table~\ref{tab:appx:diff_classif_datasets}. It is worth noting that the original ObjectNet has 313 classes, but Diffusion Classifier only uses 113 for testing. We also use a fixed noise for consistent evaluations.

\begin{table}[ht]
\centering
\caption{Dataset configurations to evaluate Diffusion Classifier.}
\label{tab:appx:diff_classif_datasets}
\begin{tabular}{@{}lcccccccc@{}}
\toprule
Dataset         & Food & CIFAR10 & Aircraft & Pets & Flowers & STL10 & ObjectNet \\ \midrule
\# classes      & 101  & 10      & 100      & 37   & 102     & 10    & 113       \\
Samples / class & 5    & 50      & 5        & 14   & 5       & 50    & 5         \\
Total samples   & 505  & 500     & 500      & 518  & 510     & 500   & 565       \\ \bottomrule
\end{tabular}
\end{table}

\section{Qualitative Results} \label{appx:more_samples}
\subsection{Open-world forgetting}
\begin{figure}[ht]
    \centering
    \includegraphics[width=1\linewidth]{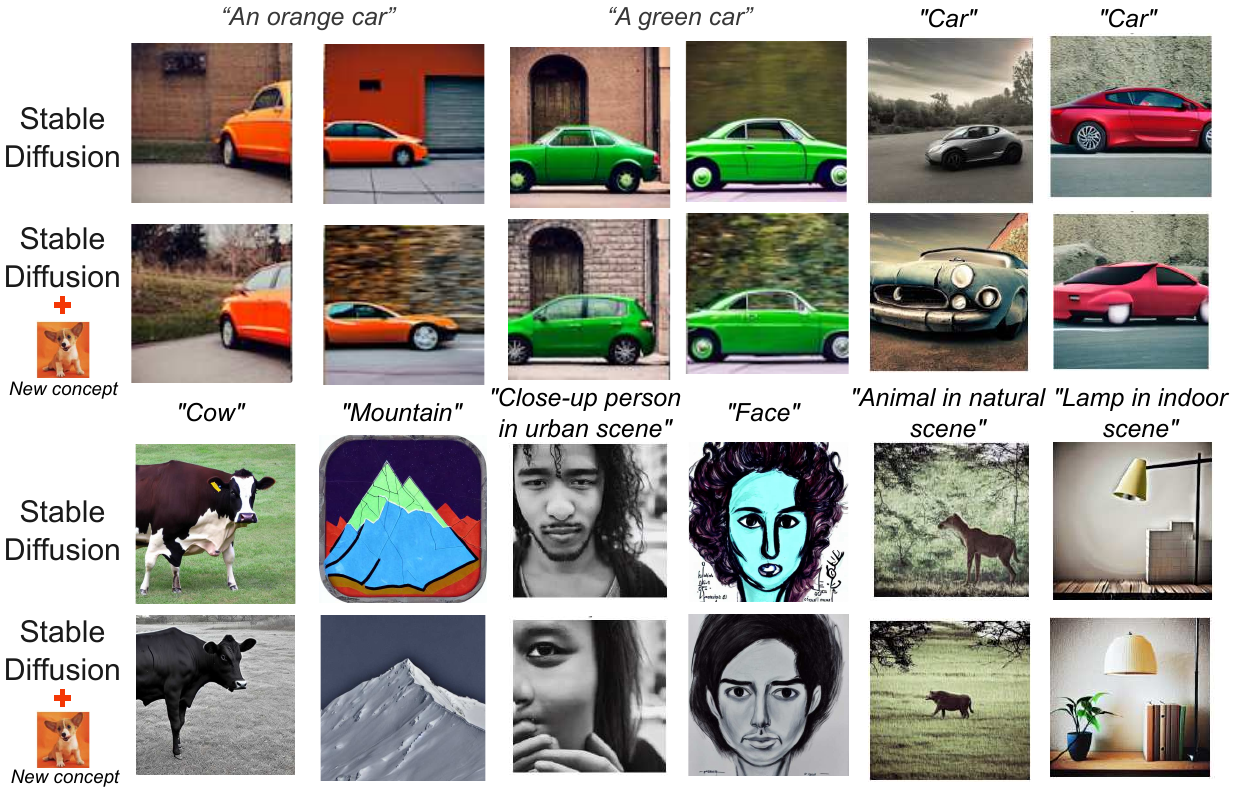}
    \caption{Several examples of appearance drift with DreamBooth. Images generated from the same initial seed. }
    \label{fig:appx:color_drift}
\end{figure}

In Figure~\ref{fig:appx:color_drift}, we present more samples of generated images with appearance drift.

\subsection{Local drift}
\begin{figure}[ht]
    \centering
    \includegraphics[width=1\linewidth]{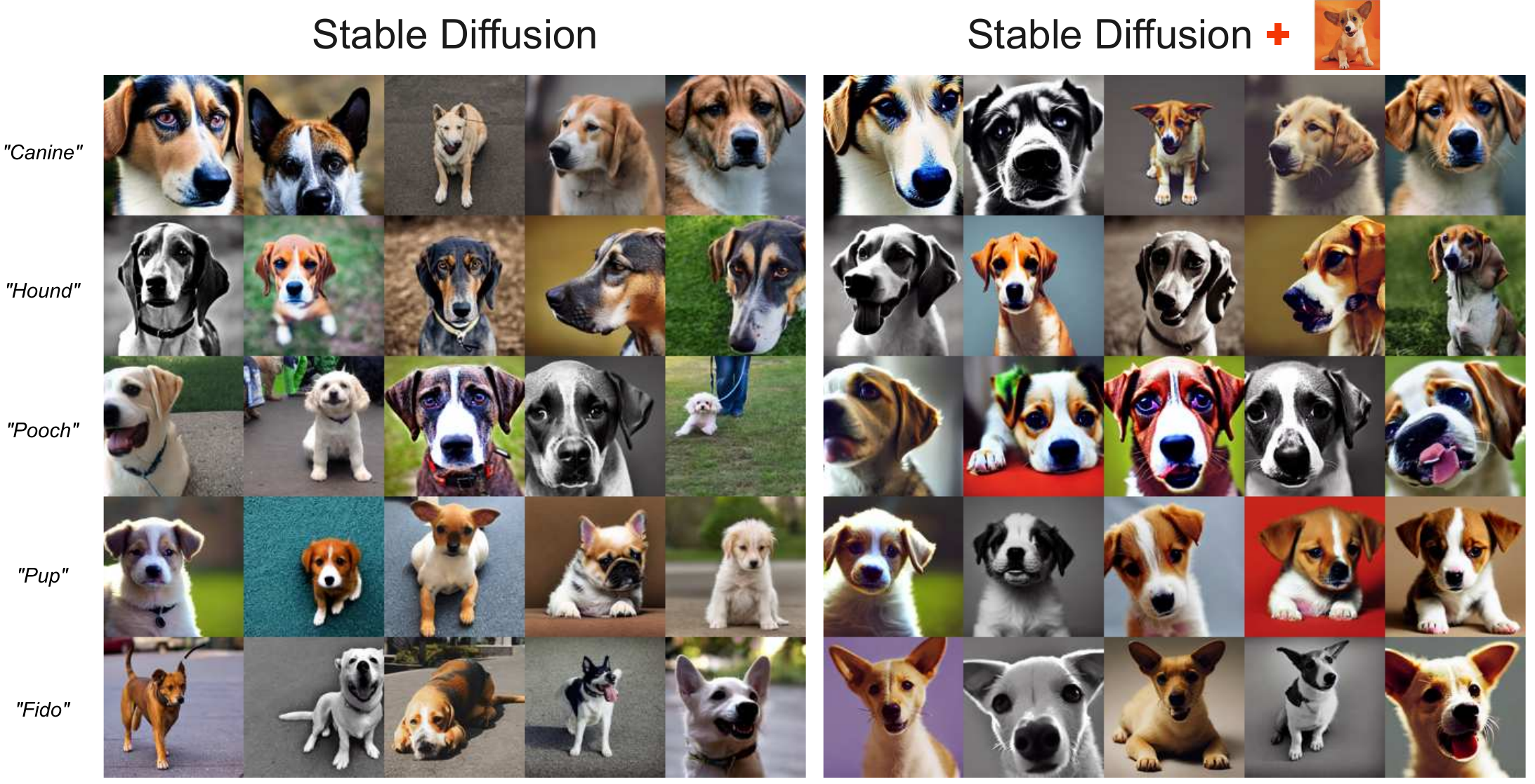}
    \caption{Several examples of local drift with DreamBooth. Images generated from the same initial seed. Note that the variety in viewpoint and breeds reduces significantly.}
    \label{fig:appx:color_drift_super}
\end{figure}

In Figure~\ref{fig:appx:color_drift_super} we present examples of local drift for the concept "dog".

\subsection{Comparison with Drift Correction}
\begin{figure}[ht]
    \centering
    \includegraphics[width=1\linewidth]{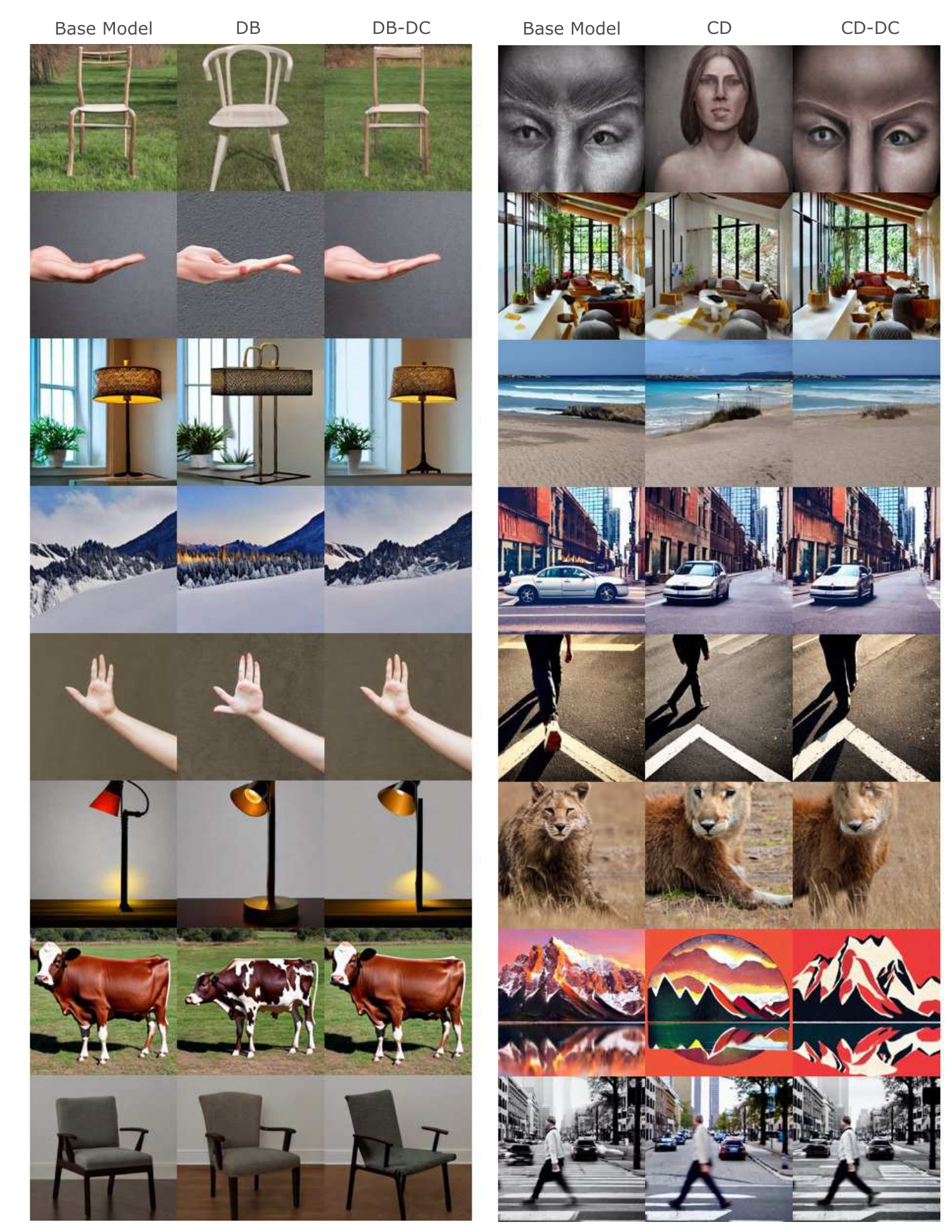}
    \caption{Qualitative result comparisons on diverse prompts for the pretrained model, a customization method and the proposed Drift Correction. These are random results and all generated from the same initial seed.}
    \label{fig:appx:qualitative}
\end{figure}

\begin{figure}[ht]
    \centering
    \includegraphics[width=1\linewidth]{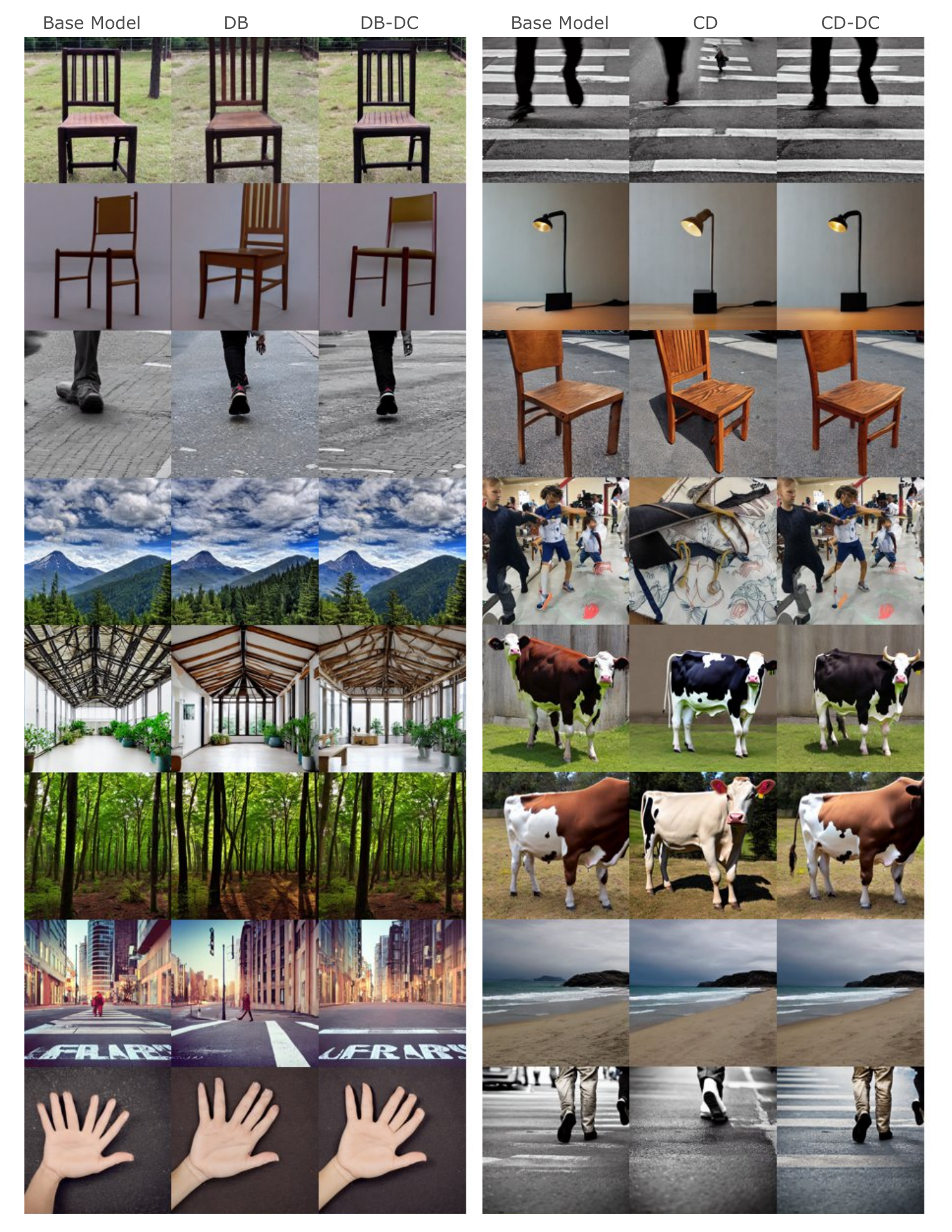}
    \caption{More qualitative result comparisons on diverse prompts for the pretrained model, a customization method and the proposed Drift Correction. These are random results and all generated from the same initial seed.}
    \label{fig:appx:qualitative2}
\end{figure}

To demonstrate the effectiveness of our proposed correction method, we provide visual examples showing how the pretrained model experiences semantic and appearance drifting, and how our method mitigates these issues. The comparative results are presented in Figures~\ref{fig:appx:qualitative} and \ref{fig:appx:qualitative2}. These examples clearly illustrate both types of drifting in the baseline model and the improvements achieved through our correction approach.

\end{document}